# BEVTraj: Map-free multimodal trajectory prediction with sensor-derived BEV context and sparse goal reasoning

Minsang Kong [1,§], Myeongjun Kim [1,§], Sang Gu Kang [2], Hejiu Lu [3], Yupeng Zhong [3] and Sang Hun Lee [3,*]

[1]Department of Automobile and IT Convergence, Kookmin University, 77 Jeongneung-ro, Seongbuk-gu, Seoul 02707, Republic of Korea
[2]Department of Automotive Engineering, Kookmin University, 77 Jeongneung-ro, Seongbuk-gu, Seoul 02707, Republic of Korea
[3]Graduate School of Automobile and Mobility, Kookmin University, 77 Jeongneung-ro, Seongbuk-gu, Seoul 02707, Republic of Korea
*Correspondence: shlee@kookmin.ac.kr
§Minsang Kong and Myeongjun Kim contributed equally.

## Abstract

Trajectory prediction is a key task in autonomous driving, where a model must forecast plausible future motions of surrounding agents from dynamic histories and scene context. Many recent methods use high-definition (HD) maps to provide explicit road topology and semantic constraints, but such maps are costly to maintain, geographically limited, and difficult to update in changing environments. This paper presents Bird's-Eye View Trajectory Prediction (BEVTraj), a map-free trajectory prediction framework that directly uses BEV features constructed from raw multi-sensor data. Since sensor-derived BEV features are dense and only implicitly structured, BEVTraj employs deformable attention to selectively aggregate agent-relevant spatial context from sparse locations rather than processing the entire BEV grid uniformly. In addition, a Sparse Goal Candidate Proposal module combines overlapping mode-wise anchor clusters with attention-based goal prediction to generate compact, scene-adaptive goal hypotheses. These goals are decoded and iteratively refined into multimodal trajectories, while trajectory-aware mode scoring and temporal consistency modeling improve confidence estimation and suppress abrupt, physically implausible refinement updates. Experiments on nuScenes and Argoverse 2 Sensor show that BEVTraj achieves competitive results under a map-free setting without using pre-built HD maps, while ablation studies and qualitative analyses illustrate the roles of BEV context aggregation and sparse goal reasoning. The source code is available at https://github.com/Kongminsang/bevtraj.

## 1. Introduction

Trajectory prediction is a key computational task in autonomous driving, where a vehicle must anticipate the future motion of surrounding traffic participants to support safe and efficient decision-making. Because future motion is uncertain and multimodal, a prediction model must reason not only from the past trajectories of agents but also from the surrounding scene context, including road geometry, free space, obstacles, and interactions with other agents. Reliable trajectory prediction therefore requires an effective representation of both dynamic motion history and static environmental structure. Many recent trajectory prediction methods rely on high-definition (HD) maps to provide structured environmental priors such as lane centerlines, road boundaries, and road topology. These maps offer compact and semantically meaningful constraints that help reduce physically implausible predictions. However, HD maps are expensive to construct and maintain, geographically limited, and difficult to update when road layouts change due to construction, accidents, temporary traffic control, or other dynamic events. As a result, prediction methods that strongly depend on pre-built HD maps may face practical limitations when deployed in unmapped or rapidly changing environments.

Existing approaches can be broadly categorized according to how they obtain and exploit environmental information, as illustrated in Figure 1. The first paradigm, pre-built HD map-based prediction, uses an existing HD map as structured environmental input for forecasting and therefore inherits the aforementioned limitations of pre-built HD maps. The second paradigm, online HD map reconstruction-based prediction, first constructs a local HD map from real-time sensor observations and then applies a map-based prediction model (Choi *et al.*, 2024; Li *et al.*, 2022; Liao *et al.*, 2022). Although this strategy reduces dependence on pre-built maps, the reconstructed map can still suffer from perception errors, missing elements, limited semantic categories, and positional inaccuracies. The third paradigm, adopted in this work, directly uses sensor-derived bird's-eye-view (BEV) features for trajectory prediction without relying on either pre-built HD maps or intermediate HD-map reconstruction. Here, *map-free* refers specifically to the absence of an explicit HD-map representation, while geometric and semantic environmental cues are still implicitly encoded in the BEV features.

Unlike vectorized HD maps, which explicitly encode lanes and road topology as sparse structured elements, sensor-derived BEV features are dense, image-like, and only implicitly structured. Directly using these features for trajectory prediction is therefore not straightforward. The model must determine which spatial regions and feature cues are relevant to a particular target agent and its possible future motion. This difficulty is amplified by the agent-centric nature of trajectory prediction: the informative context depends on the target agent's current state, candidate be-havior mode, and prediction horizon. Consequently, unlike perception tasks such as object detection, trajectory prediction is not well suited to uniform processing over the entire BEV grid. Global attention (Vaswani, 2017) and convolutional processing waste computation on largely irrelevant regions, whereas fixed local aggregation lacks the flexibility to adapt to different agents, goals, and future time steps.

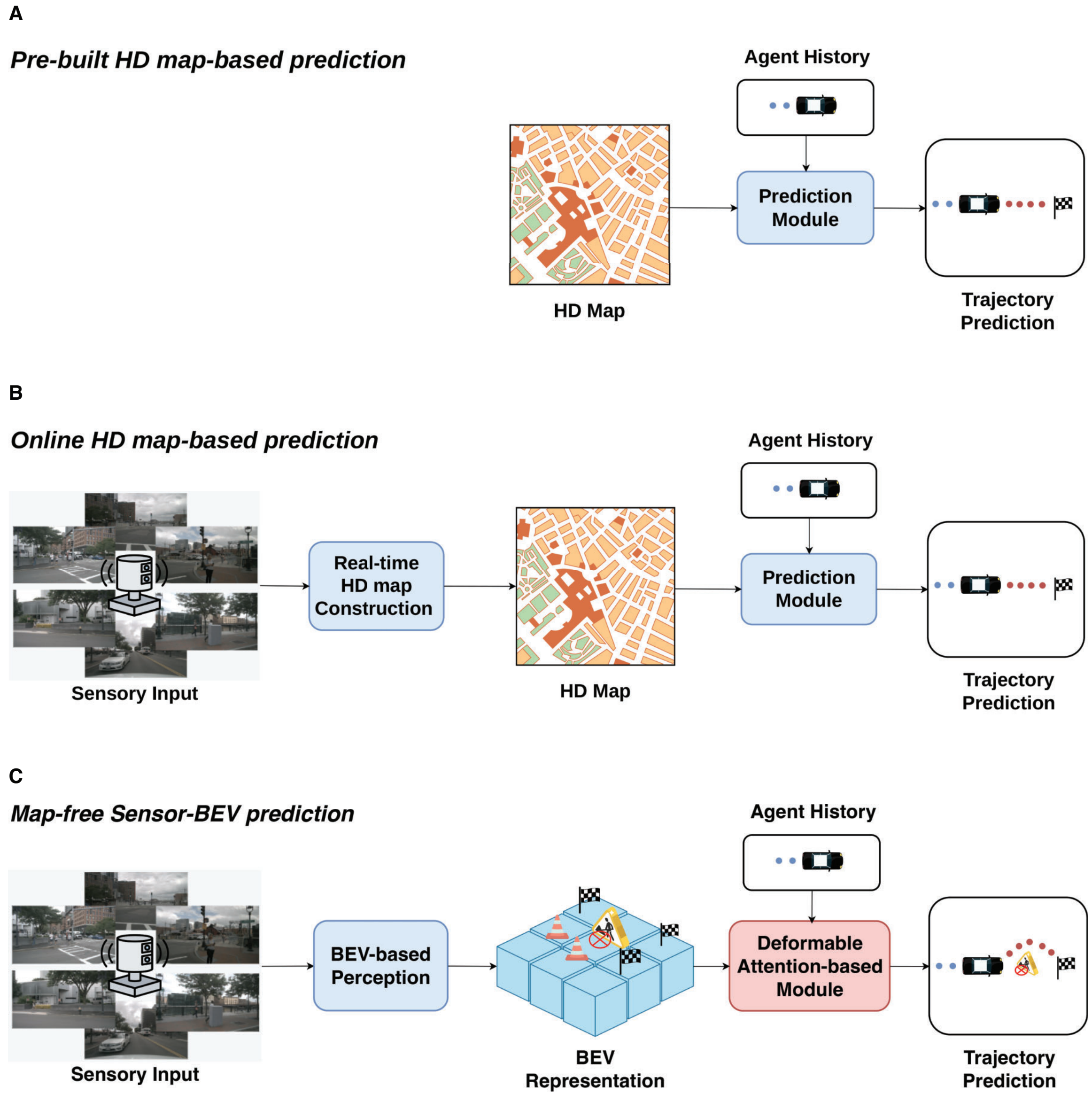

**Figure 1:** Three trajectory prediction paradigms considered in this work. (A) Pre-built HD map-based prediction uses an existing HD map as structured scene context. (B) Online HD map reconstruction-based prediction constructs a local map from real-time sensor data before prediction. (C) Map-free Sensor-BEV prediction, adopted by BEVTraj, directly uses sensor-derived BEV features without pre-built or reconstructed HD maps.

To address these challenges, we present **Bird's-Eye View Trajectory Prediction (BEVTraj)**, a map-free trajectory prediction framework that directly exploits BEV features constructed from raw multi-sensor data. BEVTraj uses deformable attention (Zhu *et al.*, 2020) to selectively aggregate task-relevant spatial context from dense BEV representations. Instead of attending to all BEV locations uniformly, the model learns a small number of query-conditioned sampling locations, allowing it to focus on scene regions that are more relevant to the target agent's future motion. In particular, for scene context encoding, we introduce *BEV Deformable Aggregation (BDA)*, which applies this deformable aggregation principle to efficiently extract relevant spatial information from the BEV feature map. This design provides an efficient way to incorporate sensor-derived spatial cues into trajectory prediction while avoiding explicit dependence on HD-map inputs.

BEVTraj further addresses multimodal prediction through a *Sparse Goal Candidate Proposal (SGCP)* module that generates a compact set of scene-adaptive goal hypotheses. Existing goal-based methods often treat densely sampled goals or predefined intention anchors directly as prediction candidates (Gu *et al.*, 2021; Shi *et al.*, 2022), which may require a large candidate set and additional selection or post-processing to achieve sufficient spatial coverage. In contrast, SGCP uses predefined anchors only as spatial references for goal reasoning rather than as the final prediction modes. Specifically, *Overlapping Mode-wise Anchor Clusters (OMAC)* associate each mode with an overlapping cluster of anchors, providing a mode-specific spatial prior while allowing neighboring modes to share plausible regions and thereby preserving scene adaptivity. *Attention-Guided Goal Prediction (AGGP)* then allows each mode query to attend to the BEV-informed anchor representations within its cluster and uses the resulting attention distribution to aggregate their spatial positions into a continuous goal prediction. This design separates semantic scene reasoning from geometric goal localization and produces a small number of adaptive goal hypotheses without directly enumerating a dense set of prediction modes.

The proposed goals are subsequently transformed into full future trajectories and iteratively refined. As refinement progressively changes each motion hypothesis, the predicted confidence should also reflect how well the latest trajectory agrees with the surrounding scene. BEVTraj therefore introduces a trajectory-aware mode scoring strategy that evaluates refined trajectories using their associated BEV context, keeping mode confidence consistent with the most up-to-date trajectory estimates. We further improve temporal consistency during refinement by suppressing abrupt and physically implausible corrections across neighboring future steps. Together, these designs enable BEVTraj to produce multimodal trajectories that are scene-consistent, temporally smooth, and physically plausible.

The main contributions of this work are summarized as follows:

- We propose BEVTraj, a map-free trajectory prediction framework that directly exploits sensor-derived BEV representations without relying on either pre-built HD maps or intermediate HD-map reconstruction.
- We develop a context aggregation strategy that selectively extracts agent- and motion-relevant spatial information from dense BEV features using deformable attention, enabling efficient and adaptive scene reasoning for trajectory prediction.
- We introduce a Sparse Goal Candidate Proposal (SGCP) module that combines OMAC with attention-based goal prediction, providing mode-specific spatial priors while preserving scene adaptivity.
- We develop an iterative refinement strategy in which future waypoints guide trajectory-aligned BEV querying, complemented by trajectory-aware mode scoring and temporal consistency modeling for reliable confidence estimation and smooth, physically plausible predictions.
- We conduct extensive experiments on nuScenes and Argoverse 2 Sensor, including quantitative comparisons, ablation studies, efficiency analyses, and qualitative evaluations, to validate BEVTraj in map-free prediction settings.

## 2. Related work

### 2.1. Trajectory prediction with HD maps

HD maps have been widely used in trajectory prediction because they provide explicit environmental constraints, including lane centerlines, road boundaries, road topology, and traffic-related semantics. Early map-based approaches often rasterize HD maps into BEV images and process them with convolutional networks (Biktairov *et al.*, 2020; Casas *et al.*, 2020, 2021; Djuric *et al.*, 2020; Marchetti *et al.*, 2020). Rasterized representations are convenient for integrating map and agent information, but they may blur lane directionality and weaken topological continuity. To preserve geometric and relational structure, later methods encode maps as vectors, polylines, or graphs. For example, VectorNet (Gao *et al.*, 2020) and LaneGCN (Liang *et al.*, 2020) model lane and agent interactions through structured vectorized representations, while LaneATT (Tabelini *et al.*, 2021) improves lane-centric feature extraction for interaction reasoning.

More recent transformer-based approaches use attention mechanisms to integrate agents with structured map elements and other heterogeneous scene information, as in Wayformer (Nayakanti *et al.*, 2023), DeMo (Zhang *et al.*, 2024), MTR (Shi *et al.*, 2022), SceneTransformer (Ngiam *et al.*, 2021), and QCNet (Zhou *et al.*, 2023). Recent extensions also explore planning- or reward-guided inference for trajectory prediction (Pei *et al.*, 2025). These methods benefit from the geometric and topological priors provided by accurate HD maps, but consequently depend on their availability, coverage, and correctness. The present work instead considers prediction directly from sensor-derived BEV features without relying on pre-built HD maps.

### 2.2. Map-free trajectory prediction

Map-free trajectory prediction aims to infer future motion without explicit HD-map inputs. This direction reduces dependence on costly map construction and maintenance, and is especially relevant for unmapped or frequently changing environments. Several map-free methods primarily rely on observed agent dynamics and inter-agent interactions. DLow (Yuan & Kitani, 2020) uses latent flows to generate diverse future trajectories, and AgentFormer (Yuan *et al.*, 2021) applies transformer-based sequence modeling to capture temporal and social interactions. Graph-based methods such as CRAT-Pred (Schmidt *et al.*, 2022) and GATraj (Cheng *et al.*, 2023) represent agents and their relations as structured graphs, while multi-level spatio-temporal models such as ML-STM (Xiang *et al.*, 2023) further strengthen interaction modeling across different temporal scales. Recent map-free approaches also investigate spatio-temporal attention and local behavior queries for vehicle trajectory prediction (Hou *et al.*, 2024; Ren *et al.*, 2025).

These methods demonstrate that meaningful trajectory prediction is possible without HD maps. Nevertheless, many map-free approaches operate mainly in trajectory or interaction space, so environmental structure such as drivable regions, road boundaries, and local scene geometry is captured only implicitly through agent motion. This can be limiting in complex scenes where future motion is strongly constrained by the surrounding layout. BEVTraj follows the map-free direction, but differs by directly using sensor-derived BEV features to provide spatial context for prediction. The goal is not to remove structure from the problem, but to replace manually curated HD-map priors with spatial cues inferred from online sensor observations. Related JCDE work has also examined multi-agent coordination in autonomous mobility systems, such as coordinated passing of multiple autonomous trucks at intersections using enhanced multi-agent reinforcement learning (Luan *et al.*, 2026).

### 2.3. BEV representation for sensor-based prediction

BEV representations provide a geometry-consistent plane in which camera, LiDAR, and other sensor information can be aligned with the trajectory prediction space. Recent BEV perception methods show that image and point-cloud features can be lifted, projected, and fused to form a common top-down representation (Huang *et al.*, 2021; Li *et al.*, 2023c; Philion & Fidler, 2020). BEVFusion (Liu *et al.*, 2023b) further demonstrates the usefulness of BEV-space sensor fusion for combining complementary semantic and geometric cues. Transformer-based BEV models such as BEVFormer (Li *et al.*, 2024), BEVFormerV2 (Yang *et al.*, 2023), and ego-centric BEV representations (Lu *et al.*, 2022) use attention mechanisms to aggregate spatial and temporal information in this representation. Deformable attention (Zhu *et al.*, 2020) is particularly relevant for dense BEV features because it samples a small set of spatial locations instead of attending to the entire feature map.

BEV representation is also compatible with sensing modalities beyond camera and LiDAR. Recent radar–camera BEV perception frameworks, such as RCBEVDet (Lin *et al.*, 2024c) and RCBEVDet++ (Lin *et al.*, 2024b), show that radar information can be lifted into BEV and fused with visual features to improve spatial completeness and dynamic cue extraction. Although the present work focuses on camera–LiDAR BEV features, these studies suggest that the BEV formulation can naturally accommodate different sensor combinations.

Recent JCDE studies further illustrate the relevance of learned visual and sensor representations in intelligent mobility systems. Kim and Lee (2026) combined transformer-based multi-view perception with intention-aware vehicle lateral control, while Hwang & Lee (2024) used three-dimensional convolutional networks for long-video-based vehicle-impact event understanding. In addition, Zhang *et al.* (2026) studied DETR-based drone object detection under adverse visual environments. Although these works address different tasks, they provide related context for sensor-driven perception and scene understanding, which motivates the use of sensor-derived BEV context in this work.

BEV-based prediction and planning frameworks show that future motion can be modeled in a sensor-derived top-down space. Some earlier approaches construct rasterized BEV inputs from HD-map geometry and process them with CNNs or MLPs (Hong *et al.*, 2019; Hou *et al.*, 2022; Li *et al.*, 2023a), but these formulations still inherit HD-map dependency. In contrast, sensor-based forecasting methods such as FIERY (Hu *et al.*, 2021), BEVerse (Zhang *et al.*, 2022b), and TBP-Former (Fang *et al.*, 2023) learn future BEV dynamics from sensor inputs and often formulate prediction as occupancy or instance-level forecasting. These approaches support the view that BEV representations can serve as useful implicit spatial priors for map-free prediction. However, many BEV forecasting methods are designed for dense occupancy-style outputs or vision-centric settings, whereas agent-centric trajectory prediction requires selecting context that depends on a target agent, its candidate goals, and future time steps. BEVTraj addresses this gap by combining sensor-derived BEV representations with agent-centric trajectory decoding, using deformable attention to selectively extract spatial context relevant to each agent's possible future motion.

## 2.4. Feature aggregation into compact queries

Attention-based architectures commonly aggregate task-relevant information from dense scene features into a compact set of query representations, reducing the amount of information that must be processed in subsequent task-specific reasoning. CenterFormer (Zhou *et al.*, 2022) uses object-center features as queries to aggregate surrounding BEV features for 3D detection, whereas PIP (Jiang *et al.*, 2022) and VAD (Jiang *et al.*, 2023) use agent and map queries to extract task-specific representations from shared BEV feature maps. Fully sparse approaches such as SparseBEV and SparseDrive further avoid explicit dense BEV construction and represent the scene directly with sparse queries (Liu *et al.*, 2023a; Sun *et al.*, 2024b). BDA similarly produces compact task-specific representations, but makes the aggregation explicitly prediction-oriented by using future-motion-conditioned spatial references to selectively extract sensor-derived features from regions relevant to possible agent motion.

## 2.5. Spatial anchors for multimodal goal prediction

Goal-based trajectory prediction first identifies plausible future destinations and then generates trajectories conditioned on them, allowing long-term multimodality to be structured through global intentions before detailed motion is decoded (Gu *et al.*, 2021; Shi *et al.*, 2022). Spatial priors for multimodal prediction have been represented in different ways. MultiPath assigns each output mode a fixed full-trajectory anchor and predicts residuals from it, whereas MTR associates each motion query with a clustered endpoint intention point (Chai *et al.*, 2019; Shi *et al.*, 2022). In both approaches, each spatial prior is tied to a dedicated trajectory hypothesis, coupling the number of represented motion modes with the number of trajectories to be decoded. DenseTNT instead estimates a dense goal distribution over lane-based candidates derived from explicit lane topology and selects a compact final goal set before trajectory completion (Gu *et al.*, 2021). In contrast, our SGCP decouples spatial goal coverage from the number of decoded trajectories by allowing each sparse mode to reason over an overlapping subset of a larger shared anchor set. This enables broad two-dimensional goal coverage with a compact set of output modes without explicit lane topology.

# 3. Method

In this section, we describe BEVTraj, a map-free trajectory prediction framework that predicts the multimodal future trajectory distribution of a target agent from sensor-derived BEV features and agent motion histories. As shown in Figure 2, BEVTraj consists of two main components: a scene context encoder and an iterative deformable decoder. The scene context encoder combines BEV features and agent motion features to construct a compact scene representation, while the iterative deformable decoder generates sparse goal candidates and refines multimodal future trajectories.

## 3.1. Scene context encoder

The Scene Context Encoder converts raw sensor observations and past agent states into a representation suitable for trajectory prediction. It consists of three parts: a sensor encoder that produces BEV features, a pre-encoder that models temporal and social information from agent histories, and a BEV Deformable Aggregation (BDA) module that summarizes dense BEV features into a compact set of spatial tokens.

### 3.1.1. *Sensor encoder*

BEVTraj uses a BEV representation because trajectory prediction is naturally defined on the ground plane and because camera and LiDAR features can be aligned in this space. In our implementation, we adopt BEVFusion (Liu *et al.*, 2023b) as an off-the-shelf sensor encoder. Given multi-view images and point clouds, the sensor encoder produces a BEV feature map $\mathbf{B} \in \mathbb{R}^{C \times H \times W}$, where C is the channel dimension and $H$ and $W$ denote the BEV grid size. The proposed trajectory prediction module is not tied to BEVFusion; another BEV-based sensor encoder can be used if it provides a compatible BEV feature map.

### 3.1.2. *Pre-encoder*

The Pre-Encoder processes the past states of agents before they are fused with BEV context. Following the general idea of temporal–social interaction modeling (Girgis *et al.*, 2021), we first apply temporal self-attention to each agent history to capture short-term motion tendencies. Social self-attention is then applied across agents at each timestep to model interactions among traffic participants. A PointNet-like encoder (Qi *et al.*, 2017) compresses the resulting representation from $\mathbb{R}^{N_a \times T_h \times d}$ into an agent feature map $\mathbf{A} \in \mathbb{R}^{N_a \times D}$, where $N_a$ is the number of agents, $T_h$ is the number of past timesteps, and $D$ is the feature dimension.

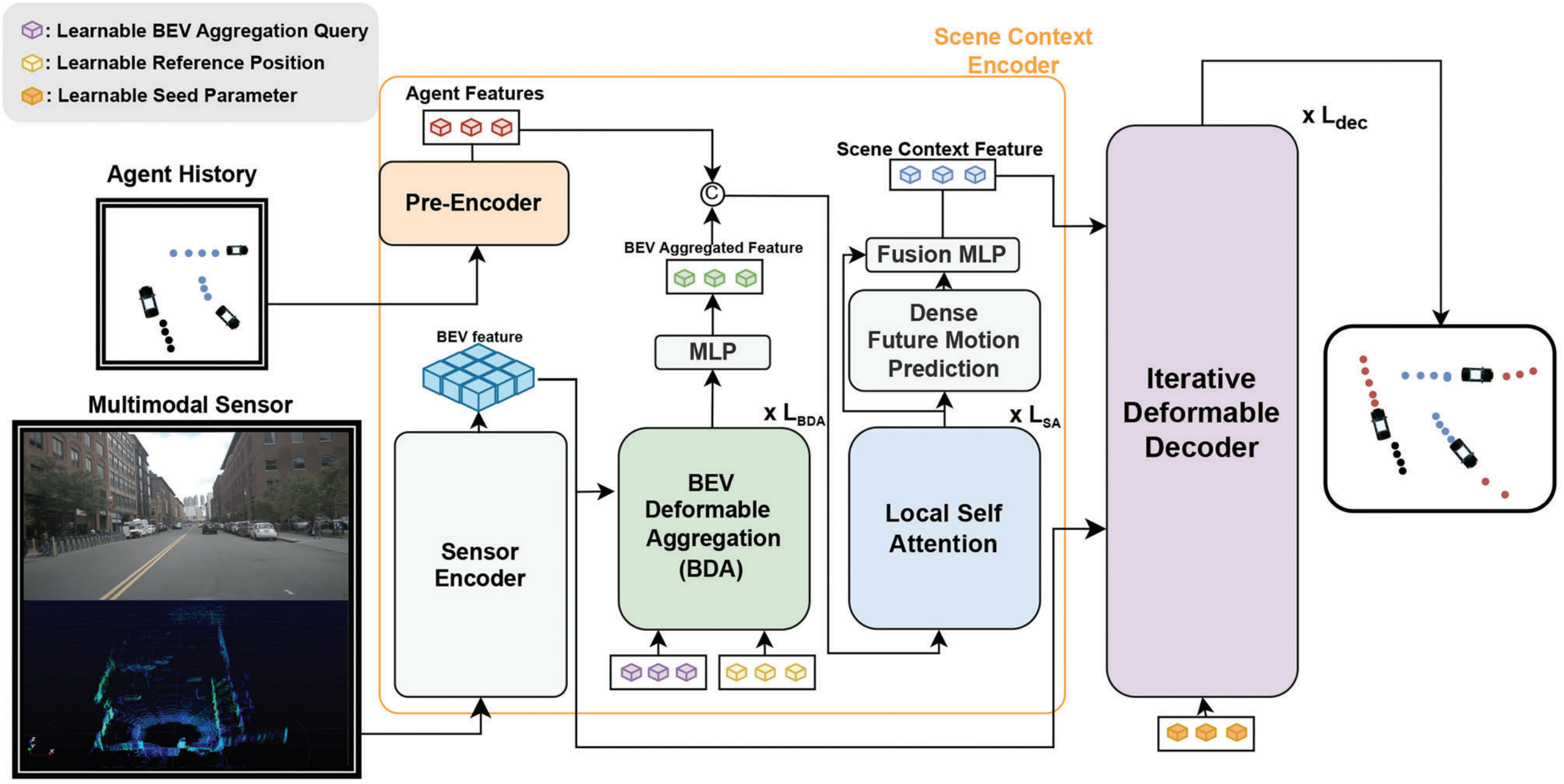

**Figure 2:** Overall architecture of BEVTraj. The sensor encoder converts multimodal sensor inputs into BEV features, and the pre-encoder represents agent motion histories. The BEV Deformable Aggregation (BDA) module summarizes relevant BEV context, which is combined with agent features to form scene context. The iterative deformable decoder then predicts and refines multimodal future trajectories for the target agent.

### 3.1.3. *BEV deformable aggregation*

Sensor-derived BEV features are dense and contain spatial information that is not uniformly relevant to trajectory prediction. Directly applying global attention to the full BEV grid is computationally expensive, whereas aggregation based on a globally fixed sampling pattern cannot explicitly adapt its spatial coverage to the current agent configuration. We therefore use a *BEV Deformable Aggregation (BDA)* module that converts the dense BEV feature map $\mathbf{B}$ into a compact set of spatially anchored BEV tokens by applying deformable attention (Zhu *et al.*, 2020) around agent-conditioned reference positions.

To construct these reference positions, we first obtain a predefined local anchor template by applying K-means clustering to future endpoints in the training set. For each scene, this template is placed around the $N_o$ agents nearest to the target agent according to their current positions and headings. This yields $N_r$ reference positions per agent and $N_{BA} = N_o N_r$ reference positions in total. Because the anchors reflect representative future-motion directions while being aligned with the current configuration of nearby agents, they provide a simple spatial prior for retrieving BEV information from regions likely to be relevant to future motion.

Since the BEV feature map $\mathbf{B}$ is represented in the ego-centric coordinate frame, reference positions defined in the target-agent-centric frame are transformed before BEV feature sampling. Let $\mathbf{r}^{\text{tar}} \in \mathbb{R}^2$ denote an arbitrary reference position in the target-agent-centric frame, $\mathbf{o}^{\text{tar}}_{\text{ego}} \in \mathbb{R}^2$ the ego-frame origin expressed in the target-agent-centric frame, and $\Delta\psi$ the relative rotation from the ego-vehicle frame to the target-agent frame. The corresponding ego-centric reference position is obtained as

$$\mathbf{r}^{\text{ego}} = \mathbf{R}(-\Delta\psi)\left(\mathbf{r}^{\text{tar}} - \mathbf{o}^{\text{tar}}_{\text{ego}}\right),$$
$$\mathbf{R}(-\Delta\psi) = \begin{bmatrix} \cos\Delta\psi & \sin\Delta\psi \\ -\sin\Delta\psi & \cos\Delta\psi \end{bmatrix}. \tag{1}$$

This transformation is used whenever target-agent-centric spatial references are employed to query the ego-centric BEV feature map, including the anchor locations used in BDA and the trajectory waypoints used throughout the iterative deformable decoder.

As illustrated in Figure 3, BDA builds on the position-aware query formulation of DAB-DETR (Liu *et al.*, 2022) and the sparse feature sampling of Deformable DETR (Zhu *et al.*, 2020). Learnable BEV Aggregate (BA) queries are associated with the reference positions, and positional embeddings derived from the reference coordinates are incorporated into the queries. Each BDA layer applies self-attention among the BA queries, followed by deformable cross-attention with $\mathbf{B}$ and a feed-forward network. The deformable cross-attention predicts sparse sampling locations around each reference position and aggregates the sampled BEV features with learned attention weights. After the BDA layers, the resulting features form the aggregated BEV representation $\mathbf{M} \in \mathbb{R}^{N_{BA} \times D}$. As shown in Figure 2, $\mathbf{M}$ is combined with the agent feature map $\mathbf{A}$ through local self-attention to model interactions between agent motion and the surrounding spatial context. Following the dense future prediction strategy of MTR (Shi *et al.*, 2022), an auxiliary dense motion prediction head is used during training to encourage interaction-aware scene encoding. The final output of the Scene Context Encoder is the scene context feature $\mathbf{C} \in \mathbb{R}^{N_a \times D}$.

## 3.2. Iterative deformable decoder

The Iterative Deformable Decoder predicts multimodal future trajectories for the target agent from the BEV feature map $\mathbf{B}$ and the scene context feature $\mathbf{C}$. As shown in Figure 4, it consists of three stages: Sparse Goal Candidate Proposal (SGCP), Initial Trajectory Prediction (ITP), and Iterative Trajectory Refinement (ITR). The decoder first proposes a compact set of candidate goals, converts them into full trajectory hypotheses, and then refines the trajectories by repeatedly querying scene-context and BEV features. The proposed modules are designed around a common principle: dense sensor-derived BEV features must be converted into sparse spatial evidence that is conditioned on the target agent, the motion hypothesis, and aligned with individual future waypoints.

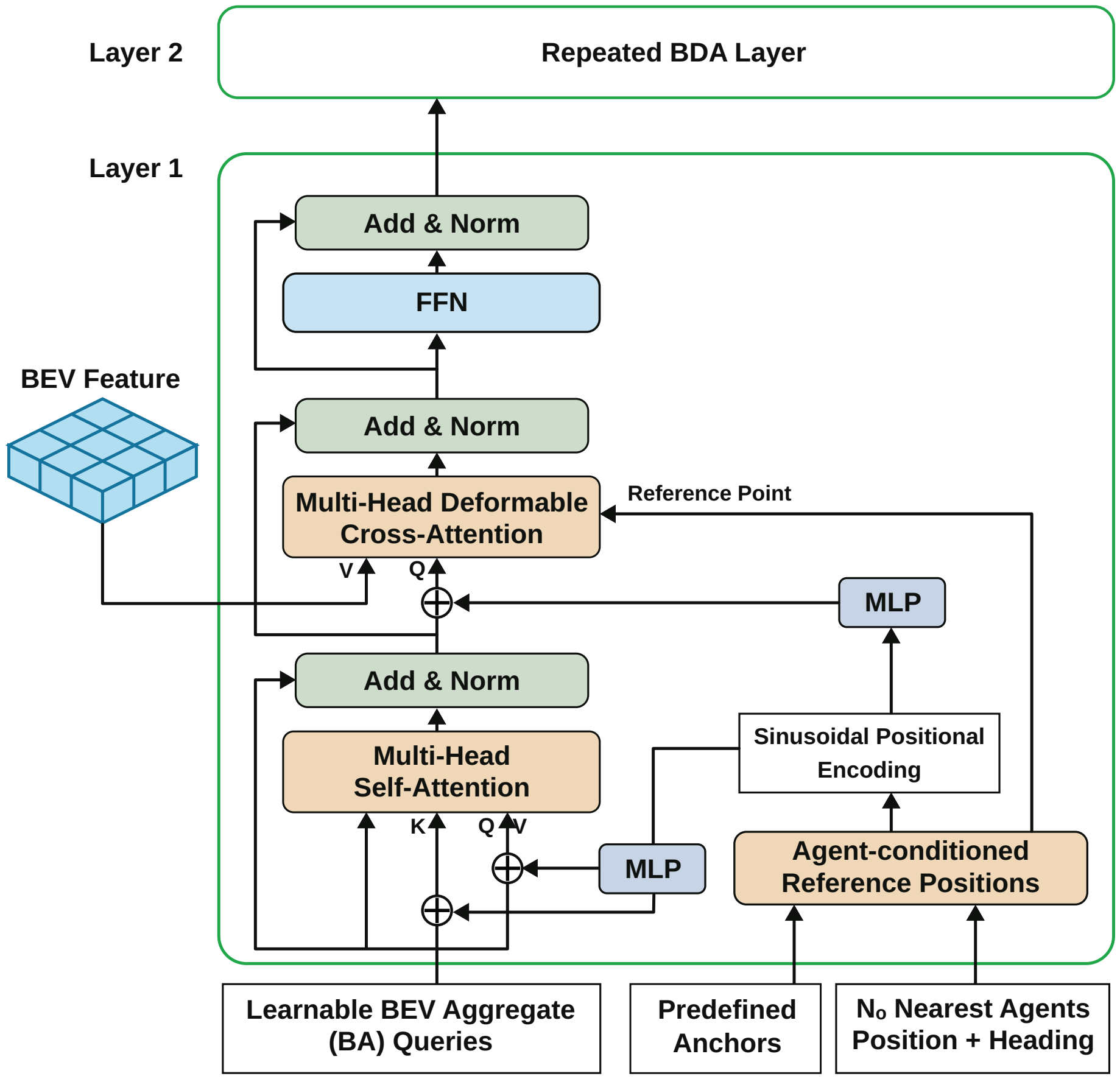

**Figure 3:** Architecture of the BEV Deformable Aggregation (BDA) module. Predefined local anchors are placed around nearby agents according to their current positions and headings to construct agent-conditioned reference positions, which guide sparse BEV feature aggregation into a compact set of spatially anchored tokens.

### 3.2.1. *Sparse goal candidate proposal*

Goal-based trajectory prediction methods often achieve broad multimodal coverage by using a large set of candidate goals (Gu *et al.*, 2021) or predefined intention anchors (Shi *et al.*, 2022). In our map-free setting, however, explicit lane topology is unavailable, making it difficult to directly construct goal candidates along lane structures as in pre-built HD-map-based approaches. Relying instead on a large set of predefined spatial candidates can preserve broad coverage, but directly associating these candidates with trajectory prediction modes increases the number of trajectory branches that must be decoded, leading to additional computational and memory costs (Shi *et al.*, 2022; Sun *et al.*, 2024a). Moreover, anchor-based methods such as MTR associate individual motion queries with predefined intention anchors and use anchor-based hard assignment to encourage mode specialization; consequently, each training sample provides positive trajectory-regression supervision only to the query associated with the matched anchor, while the remaining queries receive no positive regression update for that sample (Lin *et al.*, 2024a; Shi *et al.*, 2022). To address these limitations, SGCP decouples spatial goal exploration from the number of output trajectory modes. It uses a compact set of mode queries, while allowing each query to select its goal from a broader set of spatial candidates, thereby maintaining diverse spatial coverage without requiring a large number of trajectory prediction branches. Figure 5 provides a conceptual comparison between SGCP and representative goal candidate proposal strategies.

SGCP starts from learnable seed parameters $\mathbf{P} \in \mathbb{R}^{K \times D}$. To provide a minimal prior for goal prediction, the dynamic state history of the target agent $\mathbf{s} \in \mathbb{R}^{T_h \times K_{\mathrm{attr}}}$ is encoded and injected into the seed queries through Feature-wise Linear Modulation (FiLM) (Perez *et al.*, 2018). This conditioning encourages the predicted goal hypotheses to remain consistent with the target agent's recent motion dynamics. SGCP also employs a decoder-side BDA with a fixed set of spatial anchors for goal reasoning. Unlike the encoder-side BDA, which places local anchor templates around multiple nearby agents, the decoder-side BDA uses a target-agent-centered anchor set, where each anchor represents a potential goal region and is transformed into the ego-centric frame using Equation 1 before querying the BEV feature map. The history-conditioned seed queries are further refined through masked cross-attention with the decoder-side BDA tokens, where each query attends only to the spatial subset assigned to its mode, followed by self-attention to produce the mode queries used for goal prediction. The mode-specific subsets are defined by the OMAC described below.

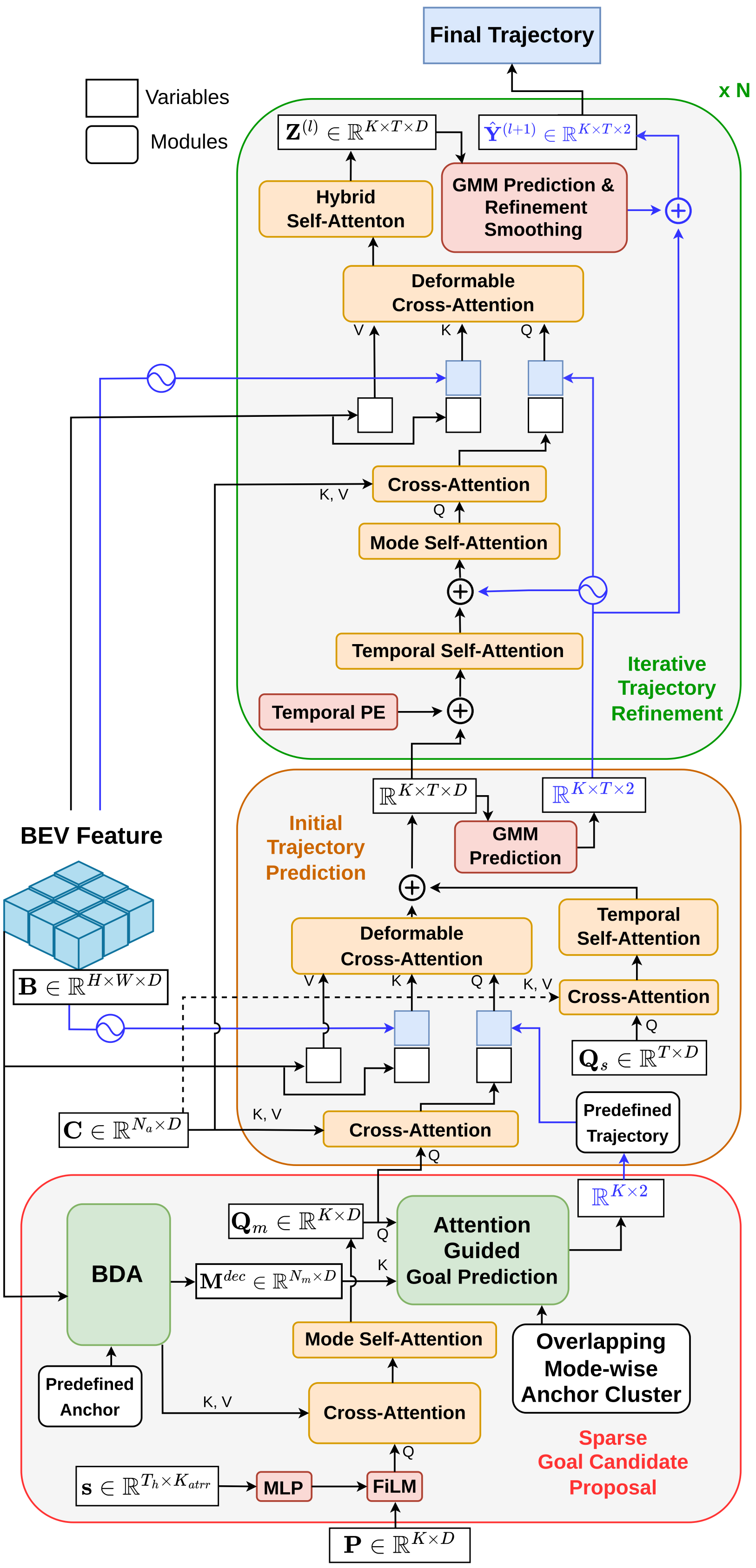

**Figure 4:** Structure of the iterative deformable decoder. Sparse goal proposal generates candidate endpoints, initial trajectory prediction converts them into trajectory hypotheses, and iterative refinement updates the trajectories using BEV and scene-context features.

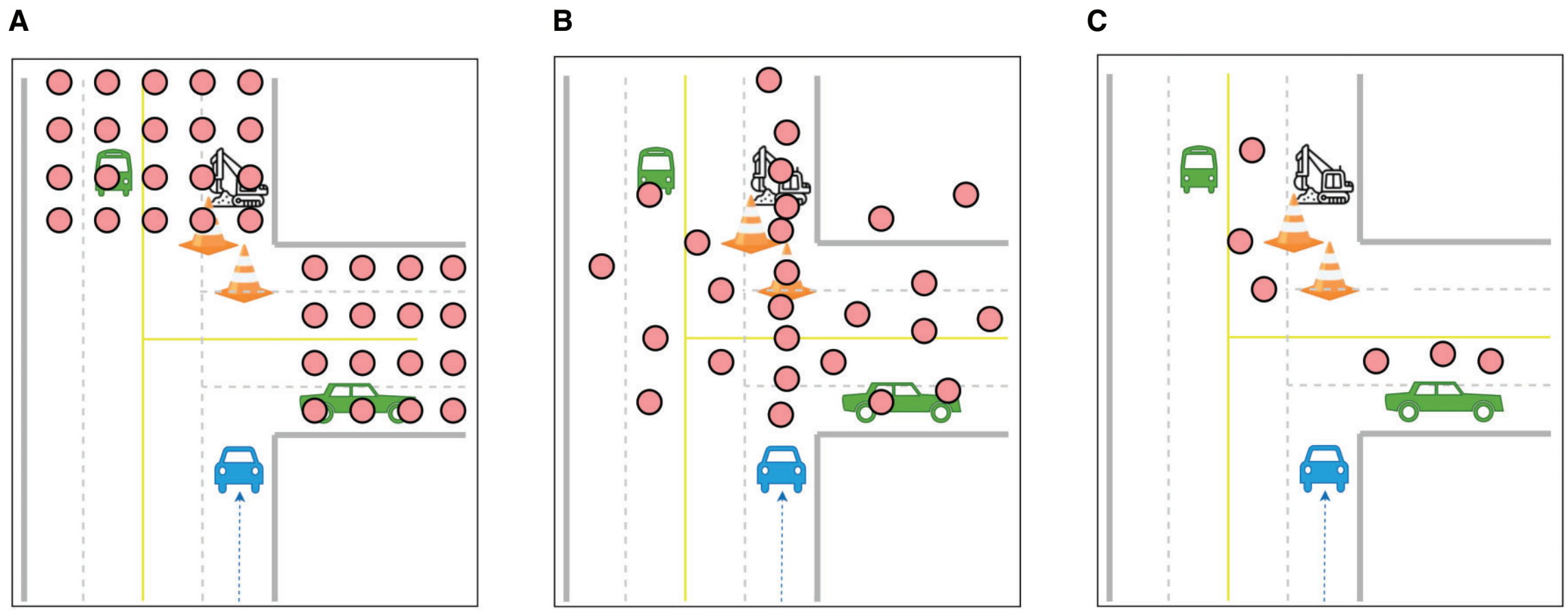

**Figure 5:** Comparison of goal candidate proposal strategies. (A) DenseTNT samples dense goals along lanes. (B) MTR uses intention points obtained by clustering training trajectories. (C) SGCP predicts a compact set of scene-adaptive goals from the target agent state and BEV context.

**Overlapping Mode-wise Anchor Cluster.** To encourage different mode queries to represent distinct motion hypotheses, we equip each mode with an explicit spatial prior. A common approach is to associate predefined spatial anchors directly with individual trajectory modes (Chai *et al.*, 2019; Shi *et al.*, 2022). However, sufficiently covering diverse motion intentions would then require a large number of trajectory modes, increasing computational and memory costs and potentially requiring additional post-processing such as non-maximum suppression (NMS) to remove redundant predictions (Shi *et al.*, 2022; Sun *et al.*, 2024a). Alternatively, dividing a larger anchor set into disjoint subsets for a small number of modes overly restricts the spatial region accessible to each mode, reducing the ability of the queries to adapt their goals to the observed scene. We therefore organize the predefined anchors into OMAC, where the $k$th mode is assigned an anchor index set $\mathcal{A}_k \subseteq \{1, \ldots, N_m\}$, with $N_m$ denoting the number of predefined anchors, and individual anchors may be shared by multiple modes, as illustrated in Figure 6.

Specifically, we first apply K-means clustering to the predefined anchor set to obtain a centroid for each mode, and define $\mathcal{A}_k$ as the indices of a fixed number of anchors nearest to the corresponding centroid. Since these nearest-neighbor sets are constructed independently for each centroid, an anchor may belong to multiple mode-wise clusters. During decoder-side BDA cross-attention, $\mathcal{A}_k$ is used as an attention mask such that the $k$-th mode query attends only to the BDA tokens associated with the anchors in its cluster. The same anchor subset subsequently defines the candidate region from which the corresponding mode query determines its scene-adaptive goal.

**Attention-guided Goal Prediction.** Directly regressing goal coordinates from anchor-conditioned BEV features requires a single prediction network to jointly interpret semantic information encoded in the BEV features and geometric information provided by the anchor locations, which can make goal regression difficult to learn. Alternatively, goal prediction can be formulated as classification over predefined goal candidates (Gu *et al.*, 2021; Zhao *et al.*, 2021); however, categorical supervision does not explicitly account for the geometric relationships among discrete goal candidates (Liu *et al.*, 2019), while selecting only the highest-probability anchor discards the probability mass over neighboring candidates and can make the prediction sensitive to an erroneously high-confidence anchor. To address these limitations, we therefore propose *attention-guided goal prediction (AGGP)*, which formulates goal prediction as attention-guided aggregation over the spatial anchors. As illustrated in Figure 7, the key idea is to let the learned features determine the compatibility of potential goals, while explicitly using the anchor coordinates to provide their geometric locations.

Formally, let $\mathbf{Q}_m = [\mathbf{q}_{m,1}, \ldots, \mathbf{q}_{m,K}]^\top \in \mathbb{R}^{K \times D}$ denote the mode queries and $\mathbf{M}^{\text{dec}} = [\mathbf{m}^{\text{dec}}_1, \ldots, \mathbf{m}^{\text{dec}}_{N_m}]^\top \in \mathbb{R}^{N_m \times D}$ denote the decoder-side BDA tokens, where $\mathbf{m}^{\text{dec}}_i \in \mathbb{R}^D$ is associated with the anchor position $\mathbf{p}^{\text{tar}}_i \in \mathbb{R}^2$. For the $k$-th mode, only the anchor-token pairs indexed by $\mathcal{A}_k$ participate in goal prediction. We first construct the goal-selection query and key as

$$\mathbf{q}^g_k = \mathbf{W}_q \mathbf{q}_{m,k}, \qquad \mathbf{k}^g_i = \mathbf{W}_k \left[\mathbf{m}^{\text{dec}}_i \,\|\, \phi(\mathbf{p}^{\text{tar}}_i)\right], \tag{2}$$

where $\phi(\mathbf{p}^{\text{tar}}_i)$ denotes the sinusoidal positional encoding of the anchor position, $\|$ denotes concatenation, and $\mathbf{W}_q$ and $\mathbf{W}_k$ are learnable linear projections into a common $D$-dimensional space. The compatibility between mode $k$ and anchor $i$ is measured by the scaled dot-product score, and the resulting scores are normalized using a low-temperature softmax to concentrate the attention weights on highly compatible anchors:

$$s^{\text{attn}}_{k,i} = \frac{(\mathbf{q}^g_k)^\top \mathbf{k}^g_i}{\sqrt{D}}, \qquad \alpha_{k,i} = \frac{\exp\left(s^{\text{attn}}_{k,i}/\tau\right)}{\sum_{j \in \mathcal{A}_k} \exp\left(s^{\text{attn}}_{k,j}/\tau\right)}, \qquad i \in \mathcal{A}_k. \tag{3}$$

The goal position of each mode is finally obtained as the attention-weighted combination of its anchor coordinates:

$$\hat{\mathbf{g}}_k = \sum_{i \in \mathcal{A}_k} \alpha_{k,i} \mathbf{p}^{\text{tar}}_i. \tag{4}$$

By aggregating the anchor coordinates according to their compatibility scores, the predicted goal can lie between discrete anchor locations and retain information from multiple plausible neighboring candidates rather than relying on a single argmax selection. In addition, each mode produces only one continuous goal from its broader anchor cluster, maintaining spatial coverage without decoding a separate trajectory for every predefined anchor or requiring redundancy-removal post-processing such as NMS.

### 3.2.2. *Initial trajectory prediction*

The Initial Trajectory Prediction (ITP) module converts the mode queries and goal hypotheses produced by SGCP into full trajectory hypotheses while integrating information from the scene context feature $\mathbf{C}$ to account for surrounding-agent interactions.

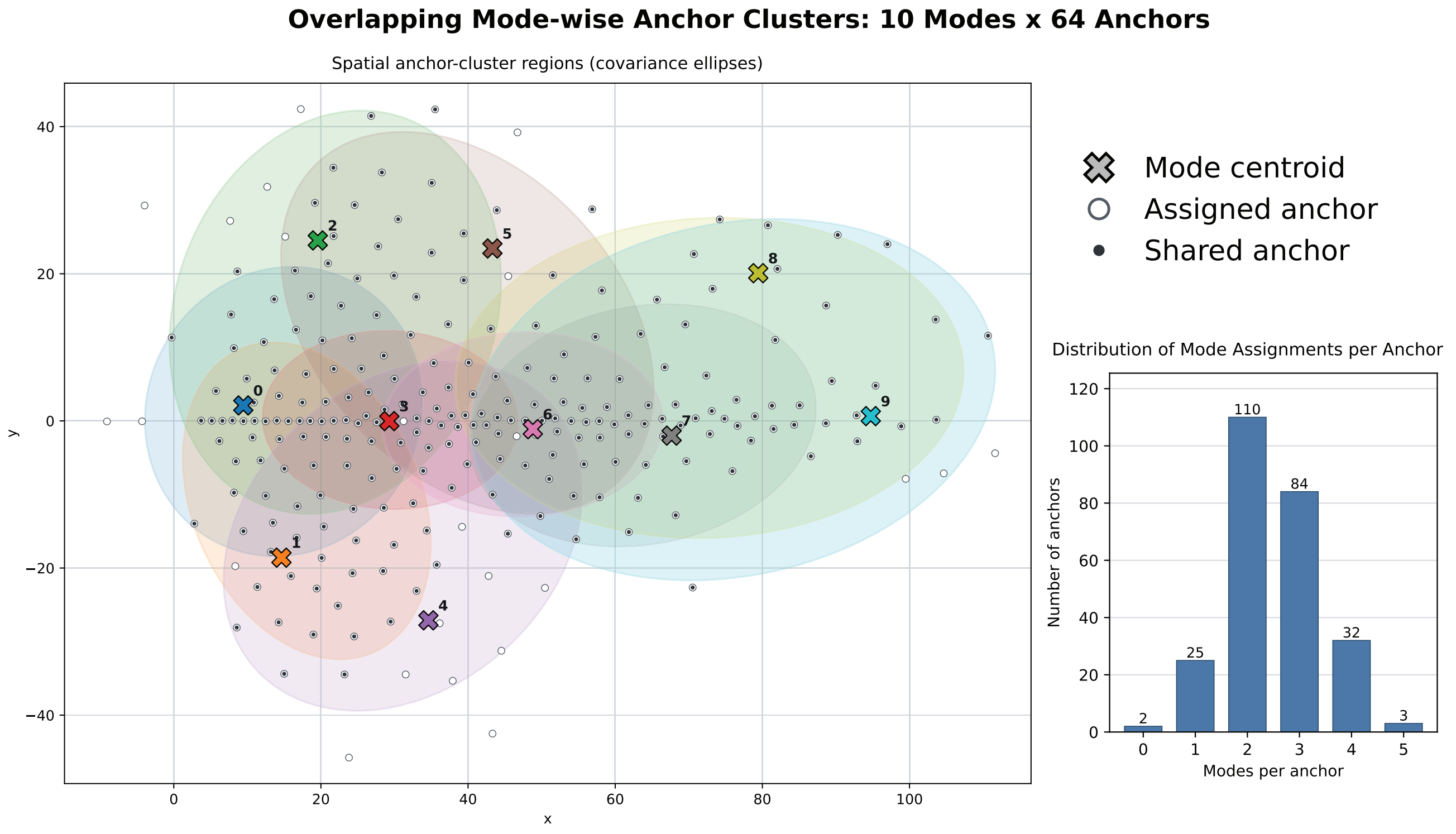

**Figure 6:** Illustration of the proposed OMAC. Each mode is associated with a centroid and a fixed-size subset of nearby anchors, while individual anchors can be shared across multiple modes. The left panel visualizes the spatial extents of the mode-wise clusters, and the right panel shows the distribution of the number of modes assigned to each anchor.

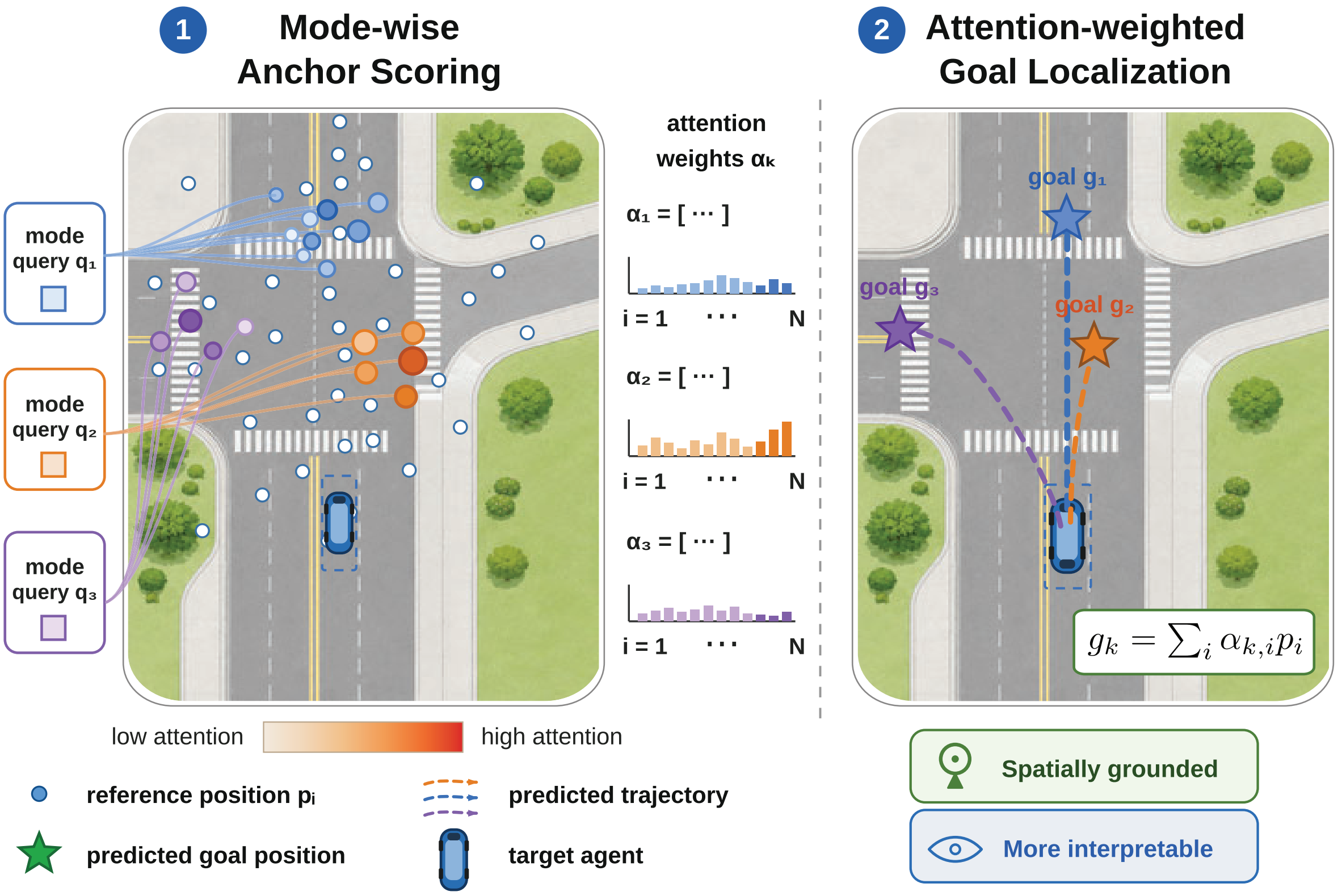

**Figure 7:** Illustration of the proposed attention-guided goal prediction. Each mode query assigns attention weights to the anchors within its mode-wise cluster according to their compatibility with the corresponding anchor-conditioned features. The final goal is obtained as an attention-weighted combination of the anchor positions, enabling continuous goal localization from a discrete set of spatial anchors.

To provide timestep-wise spatial references before trajectory decoding, each predicted goal is matched to the trajectory prototype in a predefined set whose endpoint is closest to the goal. The mode features are expanded along the prediction horizon to form timestep-wise trajectory representations in $\mathbb{R}^{K\times T\times D}$, where $T$ denotes the number of future timesteps. The waypoints of the selected trajectory prototype then serve as reference points for deformable cross-attention with the BEV feature map $\mathbf{B}$, allowing each timestep-wise representation to aggregate spatial information along a plausible path toward the predicted goal rather than only around its endpoint. In parallel, following the query-decoupling strategy of DeMo (Zhang *et al.*, 2024), a mode-independent state query $\mathbf{Q}_s \in \mathbb{R}^{T\times D}$ is used to capture shared temporal state evolution and is fused with the mode-specific trajectory representations. The resulting features are then decoded into an initial multimodal trajectory distribution using a Gaussian Mixture Model (GMM) at each future timestep (Chai *et al.*, 2019; Shi *et al.*, 2022).

### 3.2.3. *Iterative trajectory refinement*

Following the general iterative refinement paradigm used in DETR-style models (Liu *et al.*, 2022; Meng *et al.*, 2021; Zhang *et al.*, 2022a; Zhu *et al.*, 2020) and motion prediction models such as MTR (Shi *et al.*, 2022), the Iterative Trajectory Refinement (ITR) module progressively updates the initial trajectory hypotheses produced by ITP. In BEVTraj, we instantiate this paradigm at the waypoint level, with the current waypoints serving as spatial references for querying the BEV feature map $\mathbf{B}$ while incorporating the scene context feature $\mathbf{C}$. This timestep-wise design allows the spatial context to adapt along the evolving trajectory, which is particularly important for long-horizon prediction because a single reference point cannot sufficiently capture spatial dependencies over an extended motion path. Let $\mathbf{Z}^{(l)} \in \mathbb{R}^{K\times T\times D}$ and $\hat{\mathbf{Y}}^{(l)} \in \mathbb{R}^{K\times T\times 2}$ denote the timestep-wise trajectory features and trajectory estimates at refinement layer $l$, respectively, where $\mathbf{z}^{(l)}_{k,t} \in \mathbb{R}^{D}$ and $\hat{\mathbf{y}}^{(l)}_{k,t} \in \mathbb{R}^{2}$ denote their mode- and timestep-wise elements.

**Trajectory-aware Mode Scoring.** Unlike vectorized-map representations, where structured scene tokens can be repeatedly accessed during decoding, BEVTraj retrieves BEV information locally around the current trajectory hypotheses. Consequently, the spatial evidence relevant to a trajectory mode changes as its trajectory is refined. Figure 8 contrasts conventional decoder-embedding-based scoring with our trajectory-aware scoring strategy. We re-encode each refined trajectory and resample the BEV features along its updated waypoints before estimating the mode probability. This trajectory-aware scoring strategy evaluates each mode using BEV context that is spatially consistent with its latest trajectory hypothesis, rather than retaining confidence derived from an outdated trajectory representation. Note that during inference, this scoring operation is performed only once after the final refinement stage, since intermediate mode scores are not required for the final output.

**Gaussian Refinement Smoothing.** Rather than re-predicting the entire trajectory at each refinement layer, ITR predicts waypoint-wise residual offsets with respect to the previous trajectory, allowing each layer to focus on local corrections while preserving the overall motion structure. However, independently predicted offsets may introduce temporal discontinuities and trajectory jitter. We therefore apply adaptive Gaussian smoothing to the refinement offsets before updating the trajectory. For mode $k$ and timestep $t$ at refinement layer $l$, the smoothing scale and refined waypoint are computed as

$$\begin{aligned}\sigma^{(l)}_{k,t} &= \mathrm{softplus}\Big(\mathrm{Linear}\big(\mathbf{z}^{(l)}_{k,t}\big)\Big),\\ w^{(l)}_{k,t,i} &= \mathrm{softmax}_{i\in[-r,r]}\left(-\frac{i^2}{2(\sigma^{(l)}_{k,t})^2}\right),\\ \Delta\mathbf{y}^{(l),\mathrm{smooth}}_{k,t} &= \sum_{i=-r}^{r} w^{(l)}_{k,t,i}\,\Delta\mathbf{y}^{(l),\mathrm{raw}}_{k,t+i},\\ \hat{\mathbf{y}}^{(l+1)}_{k,t} &= \hat{\mathbf{y}}^{(l)}_{k,t} + \Delta\mathbf{y}^{(l),\mathrm{smooth}}_{k,t}.\end{aligned} \tag{5}$$

Since smoothing is applied only to the residual offsets, abrupt local corrections are suppressed while preserving the overall trajectory shape. An auxiliary regularization on $\sigma^{(l)}_{k,t}$ further prevents the smoothing from becoming either too weak or excessive.

## 3.3. Losses

The model is trained using supervision for goal prediction, multimodal trajectory prediction, auxiliary scene-level prediction, and trajectory regularization. The overall objective is written as

$$\mathcal{L} = \mathcal{L}_{\mathrm{goal}} + \mathcal{L}_{\mathrm{dec}} + \lambda_{\mathrm{state}}\mathcal{L}_{\mathrm{state}} + \lambda_{\mathrm{dense}}\mathcal{L}_{\mathrm{dense}} + \lambda_{\sigma}\mathcal{L}_{\sigma}, \tag{6}$$

where $\mathcal{L}_{\mathrm{goal}}$ supervises the goal hypotheses generated by SGCP, $\mathcal{L}_{\mathrm{dec}}$ trains the multimodal trajectory outputs of ITP and ITR, and the remaining terms provide auxiliary state prediction, dense future prediction, and refinement-smoothing regularization.

### 3.3.1. *Goal supervision*

Given the $K$ goal hypotheses $\{\hat{\mathbf{g}}_k\}_{k=1}^{K}$ produced by SGCP and the ground-truth trajectory $\mathbf{Y} = [\mathbf{y}_1, \ldots, \mathbf{y}_T]^\top$ with endpoint $\mathbf{g} = \mathbf{y}_T$, we first select the closest hypothesis using hard assignment:

$$k^* = \arg\min_k \|\hat{\mathbf{g}}_k - \mathbf{g}\|_2. \tag{7}$$

The selected goal is directly supervised by its endpoint distance. In addition, each goal hypothesis predicts its own final displacement error (FDE), following TPCN (Ye *et al.*, 2021). The resulting goal loss is

$$\begin{aligned}\mathcal{L}_{\mathrm{goal}} &= \lambda_{\mathrm{goal}}\|\hat{\mathbf{g}}_{k^*} - \mathbf{g}\|_2\\ &\quad + \lambda_{\mathrm{disp}}\frac{1}{K}\sum_{k=1}^{K}\mathrm{SmoothL1}\left(\hat{d}_k, \|\mathrm{sg}(\hat{\mathbf{g}}_k) - \mathbf{g}\|_2\right),\end{aligned} \tag{8}$$

where $\hat{d}_k$ denotes the predicted FDE and $\mathrm{sg}(\cdot)$ denotes stop-gradient. Thus, the FDE term trains the goal-quality estimation head without modifying the goal coordinates through its regression target.

### 3.3.2. *Multimodal trajectory supervision*

The trajectory predictions from ITP and all ITR layers are supervised using hard mode assignment. The initial positive mode is inherited from the goal assignment above, thereby preserving the structural correspondence between each SGCP goal hypothesis and the trajectory branch initialized from it. Following the evolving-anchor assignment strategy of EDA (Lin *et al.*, 2024a), the positive mode is subsequently updated at later refinement stages according to the trajectory hypotheses from preceding refinement stages. For each decoder output $l$, only the assigned mode $k^*_l$ receives trajectory-regression supervision.

Following MTR (Shi *et al.*, 2022), the predicted position distribution of the positive mode is trained using a bivariate Gaussian negative log-likelihood $\mathcal{L}^{(l)}_{\mathrm{gmm}}$. Its predicted velocity is additionally supervised with an $\ell_1$ loss $\mathcal{L}^{(l)}_{\mathrm{vel}}$, while the mode scores are trained using binary cross-entropy $\mathcal{L}^{(l)}_{\mathrm{cls}}$ with the assigned mode as the positive target. The decoder loss is

$$\mathcal{L}_{\mathrm{dec}} = \frac{1}{L}\sum_{l=1}^{L}\Big(\lambda_{\mathrm{reg}}\mathcal{L}^{(l)}_{\mathrm{gmm}} + \lambda_{\mathrm{vel}}\mathcal{L}^{(l)}_{\mathrm{vel}} + \lambda_{\mathrm{cls}}\mathcal{L}^{(l)}_{\mathrm{cls}} + \lambda_{\mathrm{acc}}\mathcal{L}^{(l)}_{\mathrm{acc}} + \lambda_{\mathrm{jerk}}\mathcal{L}^{(l)}_{\mathrm{jerk}} + \lambda_{\mathrm{curv}}\mathcal{L}^{(l)}_{\mathrm{curv}}\Big), \tag{9}$$

where $L$ denotes the number of supervised trajectory prediction outputs.

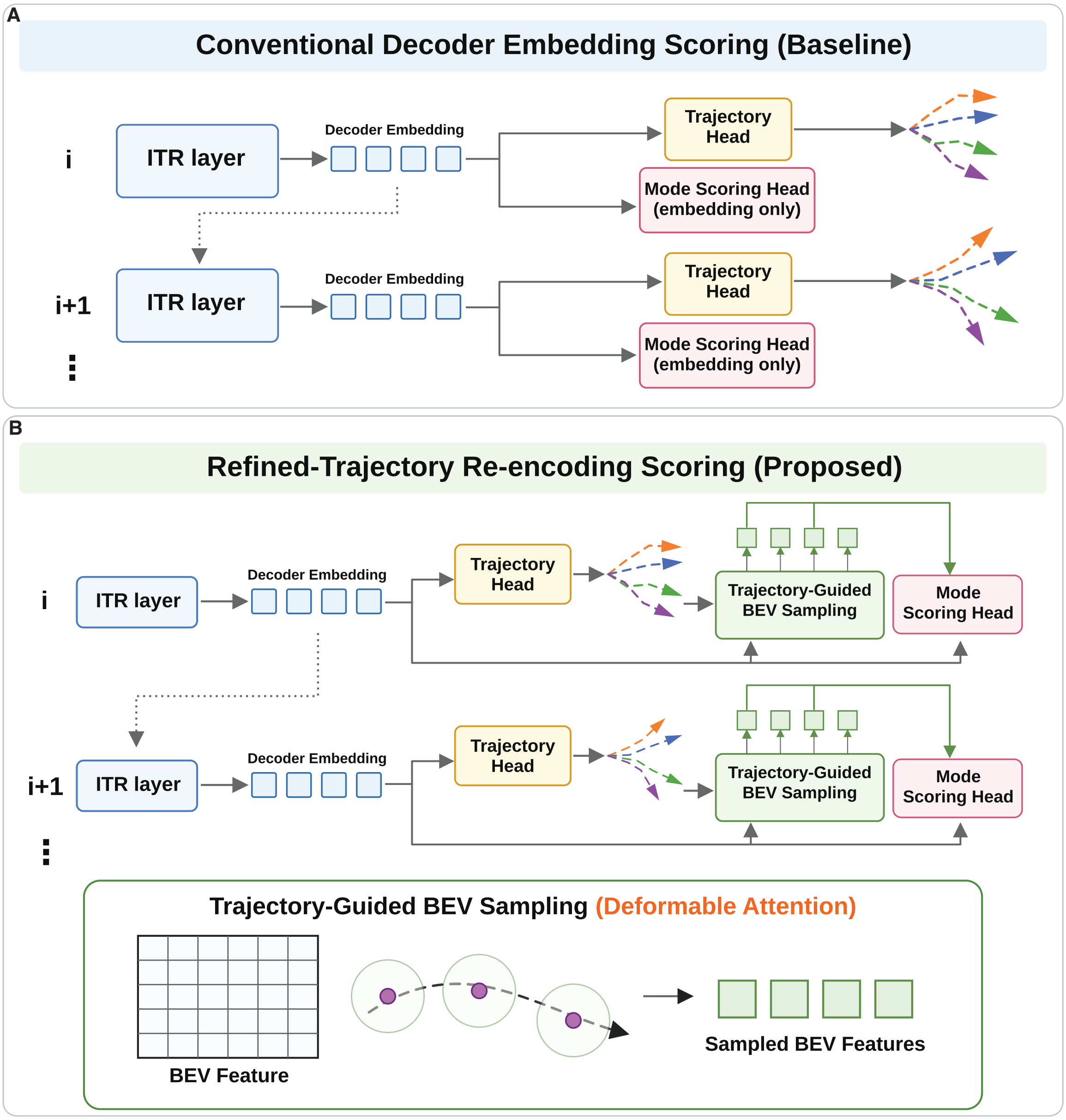

**Figure 8:** Comparison of conventional decoder-embedding-based mode scoring and the proposed trajectory-aware mode scoring. (A) Conventional scoring estimates mode confidence directly from the decoder embedding. (B) The proposed strategy re-encodes each refined trajectory and resamples BEV features along its updated waypoints using deformable attention, allowing mode confidence to be reassessed using spatial evidence consistent with the latest trajectory hypothesis.

The waypoint-wise residual updates in ITR provide fine-grained flexibility for correcting the predicted trajectory, but independently refining individual future steps can also introduce temporally inconsistent local corrections. We therefore complement the iterative refinement process with trajectory smoothness supervision that directly constrains the resulting motion sequence. In particular, penalizing acceleration, jerk, and abrupt directional changes discourages locally irregular updates from accumulating across refinement layers while preserving the ability of ITR to progressively correct the trajectory. For decoder output $l$, we denote the predicted waypoint of the assigned positive mode as $\bar{\mathbf{y}}^{(l)}_t = \hat{\mathbf{y}}^{(l)}_{k^*_l,t}$ for notational simplicity, and let $\Delta t$ denote the sampling interval. The acceleration and jerk losses are defined as

$$\mathcal{L}^{(l)}_{\mathrm{acc}} = \frac{1}{T}\sum_{t}\left\|\frac{\bar{\mathbf{y}}^{(l)}_{t+1} - 2\bar{\mathbf{y}}^{(l)}_{t} + \bar{\mathbf{y}}^{(l)}_{t-1}}{\Delta t^2}\right\|_2^2, \tag{10}$$

$$\mathcal{L}^{(l)}_{\mathrm{jerk}} = \frac{1}{T}\sum_{t}\left\|\frac{\bar{\mathbf{y}}^{(l)}_{t+2} - 3\bar{\mathbf{y}}^{(l)}_{t+1} + 3\bar{\mathbf{y}}^{(l)}_{t} - \bar{\mathbf{y}}^{(l)}_{t-1}}{\Delta t^3}\right\|_2^2. \tag{11}$$

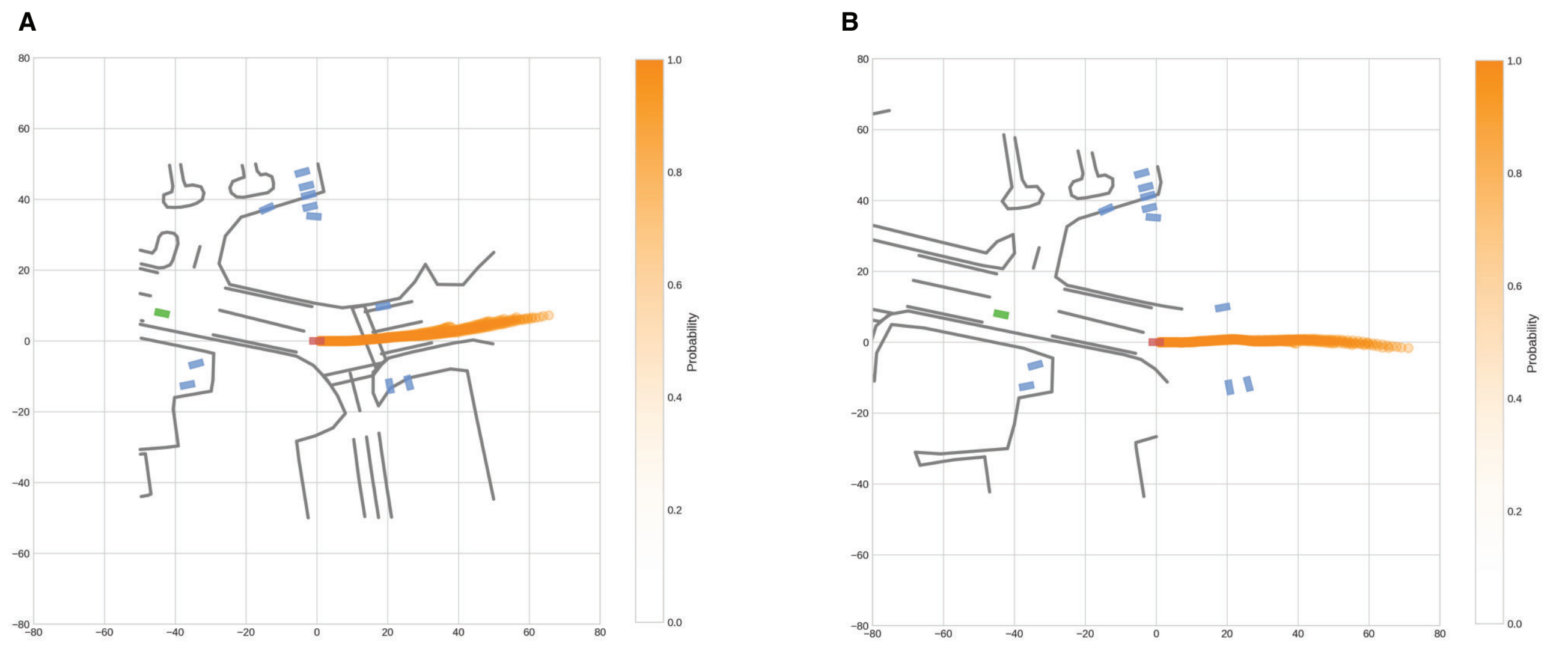

**Figure 9:** Illustration of ego-centered spatial coverage and the HD-map range settings used for baseline comparison. Red indicates the target agent, and green indicates the ego vehicle. (A) A ±50 m HD-map region centered on the target agent. (B) An ego-centered region can leave part of a distant target trajectory outside the available spatial context, illustrating the motivation for trajectory-coverage filtering; such insufficiently covered targets are excluded from the evaluation split. For the HD-map-based baselines, we use a ±50 m ego-centered map region to match the local spatial coverage of sensor-derived BEV perception.

To further suppress abrupt changes in motion direction, we define $\Delta\bar{\mathbf{y}}_t^{(l)} = \bar{\mathbf{y}}_{t+1}^{(l)} - \bar{\mathbf{y}}_t^{(l)}$ and use

$$\mathcal{L}_{\text{curv}}^{(l)} = \frac{1}{T}\sum_t \left(1 - \frac{\left(\Delta\bar{\mathbf{y}}_t^{(l)}\right)^\top \Delta\bar{\mathbf{y}}_{t+1}^{(l)}}{\left\|\Delta\bar{\mathbf{y}}_t^{(l)}\right\|_2 \left\|\Delta\bar{\mathbf{y}}_{t+1}^{(l)}\right\|_2}\right). \tag{12}$$

The term is evaluated only for displacement pairs with sufficiently large magnitudes to avoid unstable direction estimates at near-zero speeds.

### 3.3.3. *Auxiliary and regularization losses*

We use two auxiliary prediction objectives. First, the state-consistency branch is directly supervised against the ground-truth future trajectory using a SmoothL1 loss,

$$\mathcal{L}_{\text{state}} = \frac{1}{T}\sum_t \text{SmoothL1}\left(\hat{\mathbf{y}}_t^{\text{state}}, \mathbf{y}_t\right), \tag{13}$$

Second, following MTR (Shi *et al.*, 2022), the dense future prediction head for surrounding agents is supervised using a Gaussian negative log-likelihood for future positions together with an $\ell_1$ velocity regression loss:

$$\mathcal{L}_{\text{dense}} = \mathcal{L}_{\text{dense,gmm}} + \mathcal{L}_{\text{dense,vel}}. \tag{14}$$

Finally, Gaussian Refinement Smoothing predicts a smoothing scale $\sigma_{k,t}^{(l)}$ for each mode and timestep. To prevent the learned scale from deviating excessively from the nominal smoothing range, we regularize it toward the reference scale $\sigma_0$:

$$\mathcal{L}_\sigma = \frac{1}{N_\sigma}\sum_{l,k,t}\left(\sigma_{k,t}^{(l)} - \sigma_0\right)^2, \tag{15}$$

where $N_\sigma$ denotes the total number of predicted smoothing scales. This regularization stabilizes the learned temporal smoothing while retaining the flexibility to adapt the refinement offsets to different motion patterns.

The weighting coefficients balance the corresponding supervision and regularization terms. Their values and the remaining loss-related hyperparameters are provided in Section 4.2.

## 4. Experiments

This section evaluates BEVTraj under the map-free BEV trajectory prediction setting. We first describe the datasets, evaluation metrics, model implementation, and training setup. We then compare BEVTraj with representative HD-map-based baselines under matched local-map constraints, followed by complementary occupancy-style evaluation, scenario-wise analysis, qualitative visualization, and ablation studies.

## 4.1. Datasets and evaluation metrics

### 4.1.1. *nuScenes*

We evaluate BEVTraj on the *nuScenes* dataset (Caesar *et al.*, 2020), a large-scale autonomous driving dataset collected in Boston and Singapore. The dataset includes diverse urban driving scenes such as intersections, residential areas, urban streets, and industrial zones. Following the standard prediction setting, the task is to predict 6 seconds of future agent motion from 2 seconds of past trajectory history. Starting from the official prediction split, we retain only target trajectories that are sufficiently supported by the sensor-observable BEV region. As illustrated in Figure 9, because the available sensor-derived spatial context is centered on the ego vehicle, a target trajectory may extend beyond the observable region, particularly when the target lies near its boundary. To avoid evaluating predictions primarily on spatial regions for which sensor evidence is unavailable, we transform each target trajectory into the ego-centric frame defined by the current ego pose and compute its BEV coverage as the fraction of valid trajectory timesteps that fall within the predefined observable bounds of ±35 m. Targets whose coverage falls below the predefined minimum-coverage criterion of 80% are excluded. The resulting filtered split, comprising approximately 15k training and 4k validation samples, is used consistently for BEVTraj and all evaluated baselines.

### 4.1.2. *Argoverse 2 sensor*

We also evaluate the model on the *Argoverse 2 Sensor* dataset (Wilson *et al.*, 2023), a large-scale autonomous driving dataset col-lected across six U.S. cities, including Miami, Pittsburgh, and Washington, D.C. It consists of 1,000 vehicle logs, each approximately 15 seconds long, totaling about 4.2 hours of driving data. Each log provides synchronized sensor observations from two 32-beam LiDARs and nine cameras, where LiDAR and camera data are captured at 10 Hz and 20 Hz, respectively.

Unlike nuScenes, Argoverse 2 Sensor does not provide predefined target agents for the trajectory prediction setting used in this work. We therefore construct target-agent samples based on object type, motion status, trajectory validity, and proximity to the ego vehicle. As with nuScenes, we additionally retain only targets whose trajectories are sufficiently covered by the sensor-observable BEV region, so that the evaluated samples are supported by the spatial input available to the prediction model. This procedure yields approximately 17k training samples and 3k validation samples. Each sample consists of 2 seconds of past agent history and 6 seconds of future trajectory, sampled at 10 Hz, so that the temporal prediction setting is aligned with the nuScenes-based setup.

#### 4.1.3. *Metrics*

Following the official nuScenes prediction benchmark protocol, we report standard multimodal trajectory prediction metrics: minimum average displacement error (minADE), minimum final displacement error (minFDE), and Miss Rate (MR). minADE measures the average displacement error over the prediction horizon for the best-matching trajectory among the predicted modes, while minFDE measures the final displacement error of the best-matching mode. Miss Rate measures the proportion of samples for which none of the considered trajectory modes remains within 2 m of the ground-truth trajectory over the entire prediction horizon. Following common trajectory forecasting practice, these metrics are reported for different numbers of predicted modes when required by the evaluation protocol.

### 4.2. Implementation and training

#### 4.2.1. *Implementation details*

Agent histories are normalized to the target-agent coordinate system for both datasets, following Feng *et al.* (2024). For nuScenes, the 2 Hz agent history is interpolated to 10 Hz to match the temporal resolution used for Argoverse 2 Sensor. Basic image augmentations, including rotation, scaling, flipping, and cropping, are applied only to camera inputs. Point clouds and agent trajectories are not augmented. To reduce short-term occlusion effects, multi-sweep point clouds are accumulated and motion-compensated using the ego vehicle pose before being passed to the sensor encoder.

The model follows an encoder–decoder structure. The Pre-Encoder uses 2 temporal–social attention layers to encode agent motion histories. In the Scene Context Encoder, BDA places a predefined template of 64 local anchors around each of the 8 agents nearest to the target agent, yielding 512 agent-conditioned reference positions and corresponding BEV aggregate tokens. Each BDA layer employs 8-head deformable attention with 6 sampling points per head. The resulting BEV tokens are integrated with agent features through 2 local self-attention layers. The hidden dimension is set to 256 for all token representations, and the feed-forward dimension is set to 512.

The Iterative Deformable Decoder predicts $K = 10$ trajectory modes and consists of SGCP, ITP, and 3 ITR layers. SGCP employs a decoder-side BDA with 256 predefined spatial anchors, consisting of 3 deformable aggregation layers with 8 attention heads and 6 sampling points per head. Within SGCP, the history-conditioned mode queries are refined through 2 stacked attention blocks, each consisting of cluster-restricted cross-attention to the decoder-side BDA tokens, mode self-attention, and a feed-forward network, before attention-guided goal localization. For the OMAC, each mode is assigned 96 nearby anchors, with individual anchors allowed to be shared across multiple modes. The goal-attention temperature is set to $\tau = 0.25$. In ITP, each predicted goal is associated with the closest trajectory prototype from a predefined set of 256 trajectories, whose waypoints provide the initial spatial references for trajectory decoding. The deformable cross-attention modules in ITP and ITR use 8 attention heads, with 12 and 6 sampling points per head for each waypoint query, respectively. During iterative refinement, trajectory-aware mode scoring reassesses each mode using its updated trajectory and the corresponding BEV features, while Gaussian refinement smoothing is applied to waypoint-wise residual offsets using a temporal kernel size of 7. BEVTraj-specific architectural hyperparameters, including the numbers of BEV aggregate queries, mode-wise anchors, and deformable sampling points, were selected through preliminary experiments on the nuScenes validation set and subsequently fixed across all reported experiments.

For loss weighting, we use $\lambda_{goal} = 1.0$, $\lambda_{disp} = 1.0$, $\lambda_{cls} = 2.5$, $\lambda_{reg} = 4.0$, $\lambda_{vel} = 0.5$, $\lambda_{curv} = 0.05$, $\lambda_{acc} = 1 \times 10^{-3}$, $\lambda_{jerk} = 3 \times 10^{-5}$, $\lambda_{state} = 0.25$, $\lambda_{dense} = 0.5$, and $\lambda_{\sigma} = 0.3$. These coefficients were selected through preliminary hyperparameter tuning and fixed across all reported experiments. We use $\Delta t = 0.1$ s for the temporal smoothness terms and set the reference smoothing scale to $\sigma_0 = 2.5$.

#### 4.2.2. *Training details*

Before training the trajectory prediction model, the sensor encoder is pre-trained on a BEV map segmentation task. This pre-training encourages the sensor-derived BEV representation to encode static scene structure, which is important because BEVTraj does not use pre-built HD maps as input. The full trajectory prediction model is then trained end-to-end in PyTorch using AdamW (Loshchilov & Hutter, 2017) with a weight decay of 0.01.

We use a warm-up cosine learning rate schedule. The learning rate is linearly increased to $1.0 \times 10^{-4}$ during the first 3 epochs and then decayed to $1.0 \times 10^{-5}$ using cosine annealing. A dropout rate of 0.1 and layer normalization are used for regularization, and gradients are clipped to a maximum norm of 5.0. The model is trained for 15 epochs with a batch size of 8 using 4 NVIDIA RTX 4090 GPUs. All reported experiments use these fixed loss weights, and the optimization hyperparameters are kept unchanged across datasets.

### 4.3. Quantitative results

#### 4.3.1. *Comparison with trajectory prediction baselines*

Because BEVTraj follows an agent-centric trajectory prediction formulation, we compare it with both map-free trajectory-prediction methods and representative HD-map-based methods that produce the same form of agent-level trajectory outputs. The map-free baselines provide a direct comparison without explicit map information, whereas the HD-map-based baselines allow us to examine how sensor-derived BEV context compares with prediction methods that exploit explicit map priors. In contrast, sensor-based forecasting methods such as FIERY, BEVerse, and TBP-Former are primarily formulated as BEV occupancy or instance-level forecasting approaches (Hu *et al.*, 2021; Zhang *et al.*, 2022b; Fang *et al.*, 2023), and therefore differ in prediction target and output representation. We evaluate this class of methods sep-arately through the complementary occupancy-style analysis described below.

**Table 1:** Trajectory prediction results on the *nuScenes* validation set under the matched local-input setting. Bold values indicate the best (lowest) performance for each metric.

| Method | $minADE_5$ ↓ | $minADE_{10}$ ↓ | $minFDE_1$ ↓ | $MR_5$ ↓ | $MR_{10}$ ↓ |
|---|---|---|---|---|---|
| CRAT-Pred | 2.2943 | 1.7746 | 9.9582 | 0.8648 | 0.7405 |
| GATraj | 1.8603 | 1.4287 | 10.8761 | 0.6990 | 0.5590 |
| Autobot | 1.6752 | 1.0231 | 8.4280 | 0.6499 | 0.4651 |
| MTR | 1.2072 | 1.0507 | 6.8757 | 0.5815 | 0.5578 |
| Wayformer | 1.3569 | 0.9004 | 7.0110 | 0.5411 | 0.4154 |
| DeMo | 1.2764 | **0.8596** | **5.7980** | 0.4775 | **0.3295** |
| BEVTraj (Ours) | **1.0543** | 0.8708 | 6.7496 | **0.4745** | 0.4207 |

**Table 2:** Trajectory prediction results on the *Argoverse 2 Sensor* validation set under the matched local-input setting. Bold values indicate the best (lowest) performance for each metric.

| Method | $minADE_5$ ↓ | $minADE_{10}$ ↓ | $minFDE_1$ ↓ | $MR_5$ ↓ | $MR_{10}$ ↓ |
|---|---|---|---|---|---|
| CRAT-Pred | 1.4918 | 1.1143 | 6.2778 | 0.5833 | 0.3828 |
| GATraj | 0.8217 | 0.6665 | 5.1964 | 0.3775 | 0.2797 |
| Autobot | 1.1174 | 0.6436 | 5.7746 | 0.3677 | 0.2482 |
| MTR | 0.8638 | 0.7755 | 5.0727 | 0.3843 | 0.3606 |
| Wayformer | 0.8367 | 0.5583 | 4.7865 | 0.3280 | 0.2145 |
| DeMo | **0.7656** | **0.4663** | **3.8452** | **0.2926** | **0.1408** |
| BEVTraj (Ours) | 0.8321 | 0.6786 | 5.5122 | 0.3513 | 0.2889 |

Tables 1 and 2 summarize the results on nuScenes and Argoverse 2 Sensor, respectively. Overall, BEVTraj achieves competitive performance against HD-map-based baselines while relying solely on sensor observations. Unlike pre-built maps, which provide explicit road geometry and topology throughout the input region, sensor observations are affected by occlusion, range- and distance-dependent degradation, and adverse sensing conditions, while BEVTraj must infer motion-relevant scene structure from implicit sensor evidence. The competitive performance achieved despite these observation-dependent limitations highlights the effectiveness of the proposed map-free BEV formulation. Although incorporating pre-built maps is beyond the scope of this work, these results suggest that complementing sensor-derived evidence with explicit map priors could yield further improvements.

In the main comparisons, the HD-map inputs of the map-based baselines are restricted to an ego-centered ±50 m region, as illustrated in Figure 9. This setting is chosen to provide a comparable local spatial range to the sensor-derived BEV input used by BEVTraj, which is inherently limited to ego-centered observations rather than access to unrestricted pre-built map context. To examine the effect of this spatial restriction, we additionally evaluate the map-based baselines with ego-centered map ranges of ±30, ±50, and ±100 m, as reported in Tables 3 and 4. For this auxiliary range study, we use the original, unfiltered validation samples without constructing a new split or applying the sensor-observable BEV coverage filtering used for the main results. Although broader map context improves several metrics for some methods, the gains are neither uniform nor monotonic across models and metrics, and the ±50 m setting remains comparable to the ±100 m setting in many cases. These results indicate that restricting the input range to ±50 m does not simply cripple the map-based baselines; rather, it removes access to long-range map information that is unavailable to sensor-based BEV perception while retaining substantial local scene context. We therefore use ±50 m as a controlled comparison point for matching spatial coverage, rather than as a representation of the unrestricted capability of pre-built HD maps.

### 4.3.2. *Complementary occupancy-style evaluation*

Although the primary output of BEVTraj is a set of agent-level future trajectories, we additionally provide an occupancy-style (Hu *et al.*, 2021) evaluation to examine whether the predicted trajectories are spatially coherent at the scene level. This complementary evaluation is necessary for comparison with sensor-only prediction and planning frameworks, which are predominantly formulated in terms of future BEV occupancy rather than agent-level trajectories. For this analysis, the predicted multi-agent trajectories are deterministically converted into BEV occupancy grids without using future ground-truth information. In the end-to-end setting, agent histories are constructed from 3D detections produced by LargeKernel3D (Chen *et al.*, 2023) and temporally associated using OptiPMB (Ding *et al.*, 2025), whereas the GT-history setting uses ground-truth agent histories.

As shown in Table 5, BEVTraj (E2E), which is directly comparable to the occupancy-based methods, achieves an IoU of 52.7, slightly exceeding the strongest compared occupancy-based method. The BEVTraj (GT-hist) result is reported only as an upper-bound diagnostic to isolate the effect of upstream detection and tracking errors and should not be directly compared with the end-to-end occupancy forecasting methods. This evaluation should nevertheless be interpreted as a complementary analysis rather than a direct replacement for trajectory metrics, since BEVTraj is not trained with occupancy supervision. Even so, the end-to-end result suggests that the predicted trajectories preserve a degree of scene-level spatial consistency.

## 4.4. Qualitative results

We provide qualitative analyses to complement the quantitative evaluation by examining both the internal reasoning process of BEVTraj and the resulting trajectory predictions. The vi-sualizations trace how sensor-derived BEV representations are transformed into scene-aware motion features, multimodal goal and trajectory hypotheses, and progressively refined predictions through the proposed decoder. We further examine the resulting prediction behavior across representative driving scenarios and challenging visibility conditions.

**Table 3:** Effect of the ego-centered HD-map input range on map-based baselines using the unfiltered *nuScenes* dataset. The trajectory-coverage filtering used in the main comparison is not applied. Numbers in parentheses denote the map half-range in meters.

| Method | $\text{minADE}_5 \downarrow$ | $\text{minADE}_{10} \downarrow$ | $\text{minFDE}_1 \downarrow$ | $\text{MR}_5 \downarrow$ | $\text{MR}_{10} \downarrow$ |
|---|---|---|---|---|---|
| Autobot (30) | 1.9407 | 1.1969 | 9.9123 | 0.7264 | 0.5321 |
| Autobot (50) | 1.9566 | 1.1649 | 8.8171 | 0.7302 | 0.5203 |
| Autobot (100) | 1.8475 | 1.1100 | 9.0446 | 0.6991 | 0.4994 |
| MTR (30) | 1.3228 | 1.0547 | 7.2409 | 0.6232 | 0.5604 |
| MTR (50) | 1.2926 | 1.0446 | 7.2689 | 0.6121 | 0.5553 |
| MTR (100) | 1.2722 | 1.0332 | 7.3652 | 0.6054 | 0.5502 |
| Wayformer (30) | 1.3983 | 1.0226 | 8.0459 | 0.5532 | 0.4641 |
| Wayformer (50) | 1.4034 | 0.9877 | 7.9707 | 0.5554 | 0.4503 |
| Wayformer (100) | 1.4166 | 0.9620 | 7.6164 | 0.5591 | 0.4402 |
| DeMo (30) | 1.3669 | 0.9575 | 6.9147 | 0.5031 | 0.3624 |
| DeMo (50) | 1.3137 | 1.0424 | 7.1679 | 0.4883 | 0.3892 |
| DeMo (100) | 1.3518 | 0.9254 | 6.9370 | 0.4982 | 0.3513 |

**Table 4:** Effect of the ego-centered HD-map input range on map-based baselines using the unfiltered *Argoverse 2 Sensor* dataset. The trajectory-coverage filtering used in the main comparison is not applied. Numbers in parentheses denote the map half-range in meters.

| Method | $\text{minADE}_5 \downarrow$ | $\text{minADE}_{10} \downarrow$ | $\text{minFDE}_1 \downarrow$ | $\text{MR}_5 \downarrow$ | $\text{MR}_{10} \downarrow$ |
|---|---|---|---|---|---|
| Autobot (30) | 1.3426 | 0.8128 | 7.5449 | 0.4843 | 0.3523 |
| Autobot (50) | 1.3284 | 0.7430 | 6.9975 | 0.4766 | 0.3079 |
| Autobot (100) | 1.3052 | 0.7370 | 6.8554 | 0.4642 | 0.3041 |
| MTR (30) | 0.9558 | 0.8055 | 5.8315 | 0.4473 | 0.3817 |
| MTR (50) | 0.9243 | 0.7944 | 5.8264 | 0.4254 | 0.3739 |
| MTR (100) | 0.9299 | 0.7961 | 5.9988 | 0.4292 | 0.3751 |
| Wayformer (30) | 1.0499 | 0.7009 | 6.2962 | 0.4610 | 0.3017 |
| Wayformer (50) | 1.0454 | 0.6721 | 6.2511 | 0.4581 | 0.2833 |
| Wayformer (100) | 1.0268 | 0.7122 | 6.0104 | 0.4459 | 0.3091 |
| DeMo (30) | 0.9142 | 0.6046 | 5.8348 | 0.3818 | 0.2079 |
| DeMo (50) | 0.8528 | 0.6426 | 5.2801 | 0.3440 | 0.2278 |
| DeMo (100) | 0.8133 | 0.5107 | 4.9291 | 0.3204 | 0.1614 |

**Table 5:** Complementary occupancy-style evaluation over a 100 × 100 m range. BEVTraj does not directly predict occupancy; its occupancy maps are obtained by deterministic conversion of predicted trajectories into BEV occupancy grids. BEVTraj (E2E) is the directly comparable setting, whereas BEVTraj (GT-hist) is included only as an upper-bound diagnostic using ground-truth agent histories. Bold values indicate the best (lowest) performance.

| Method | IoU ↑ |
|---|---|
| FIERY (Hu *et al.*, 2021) | 36.7 |
| StretchBEV (Akan & Güney, 2022) | 37.1 |
| ST-P3 (Hu *et al.*, 2022) | 38.9 |
| BEVerse (Zhang *et al.*, 2022b) | 40.9 |
| PowerBEV (Li *et al.*, 2023b) | 39.3 |
| UniAD (Hu *et al.*, 2023) | 40.2 |
| FusionAD (Ye *et al.*, 2023) | 51.5 |
| BEVTraj (GT-hist; upper-bound diagnostic) | 78.2 |
| BEVTraj (E2E) | **52.7** |

#### *4.4.1. Multimodal trajectory prediction across diverse scenarios*

Figure 10 presents qualitative trajectory predictions across diverse driving scenarios, including curved roads, turning maneuvers, and cluttered intersections. Across these scenes, BEVTraj produces multimodal trajectories that remain consistent with the surrounding road layout and plausible directions of travel. This behavior is particularly evident in geometrically constrained situations, where the high-probability modes tend to follow curved road structures or feasible branches through intersections rather than deviating toward spatially implausible regions.

Vectorized HD-map-based approaches must encode road geometry and semantic information into a latent representation and jointly infer how these heterogeneous cues constrain future motion, which can become challenging in scenes with complex spatial and semantic relationships. BEVTraj reduces this burden by separating semantic compatibility from geometric localization during goal prediction: its attention-guided goal prediction uses BEV features to assess plausible regions while the corresponding anchor coordinates explicitly provide spatial structure, allowing the final goal to be adjusted continuously through attention-weighted localization. Afterward, trajectory refinement is performed waypoint-wise using local BEV features around the current trajectory, so the spatial evidence used for correction changes along the predicted path rather than being summarized once for the entire scene. This local-ized and trajectory-adaptive reasoning is particularly beneficial in curved roads and complex intersections, where different portions of a future trajectory are constrained by different local scene geometries.

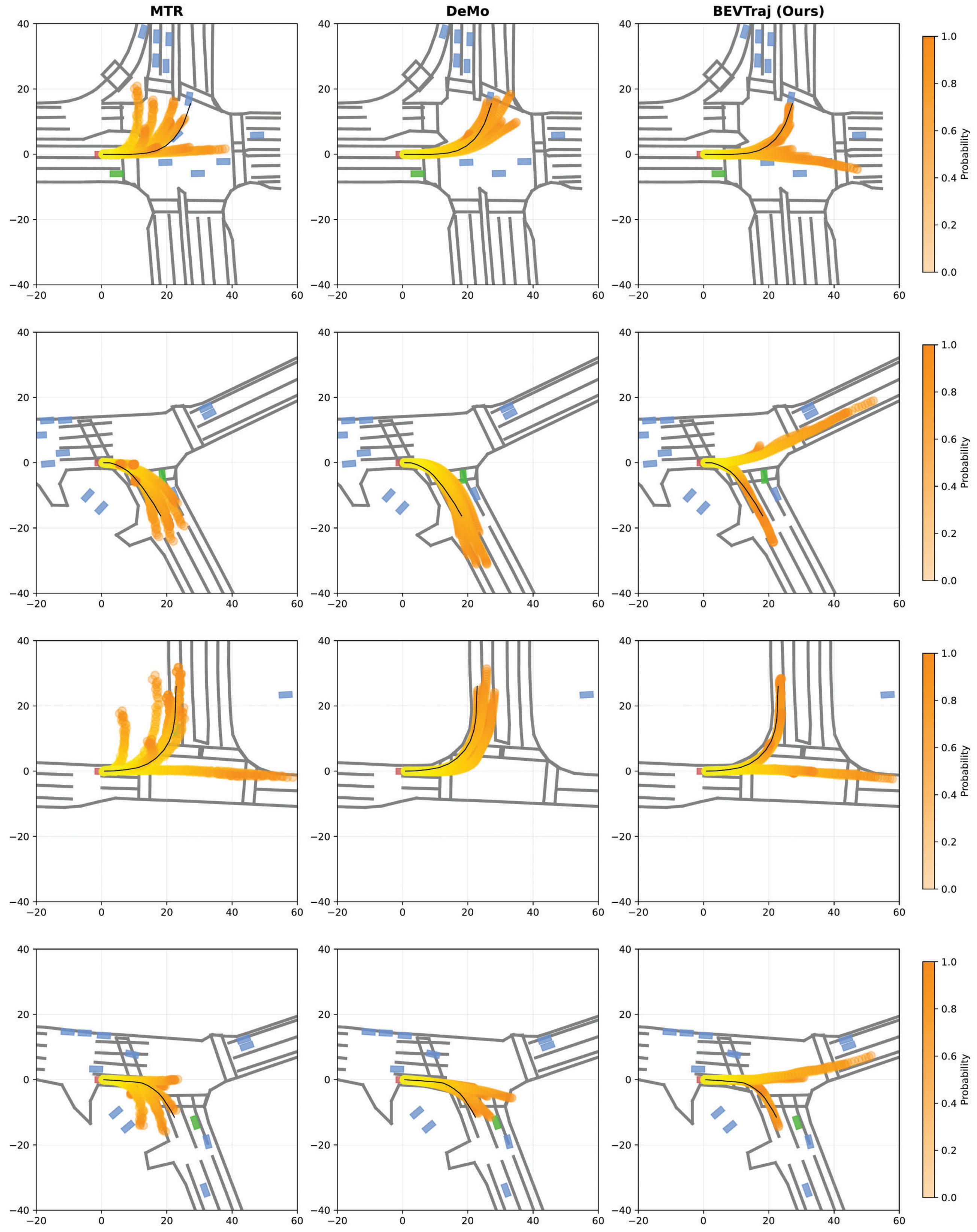

**Figure 10:** Qualitative trajectory prediction examples across diverse driving scenarios, including curved roads, turning maneuvers, and cluttered intersections. The red and green rectangles denote the target agent and the ego vehicle, respectively, while the black lines represent the ground-truth trajectories. Greater color intensity of the predicted trajectories indicates higher mode probabilities. For BEVTraj, the HD map is shown only for visualization and is not used as model input.

To further illustrate how this spatial reasoning translates into multimodal predictions, Figure 11 visualizes the goals produced by SGCP together with their initial and iteratively refined trajectories. Through OMAC, different modes retain distinct spatial priors while sharing plausible regions, enabling the predicted goals to capture diverse motion intentions without sacrificing scene adaptivity. Each goal remains associated with a corresponding trajectory branch throughout decoding, while iterative refinement adjusts the waypoint sequence according to local scene geometry without collapsing the distinct motion hypothesis represented by each mode. These examples demonstrate how spatially differentiated goals are progressively transformed into scene-adaptive multimodal trajectories.

### 4.4.2. *Visualization of BEV feature representations*

To examine the information available before trajectory decoding, we visualize intermediate BEV features constructed from camera, LiDAR, and their fusion in Figure 12. Camera-derived BEV features tend to contain semantic and contextual cues from the visual scene, while LiDAR-derived features provide sharper geometric structure related to obstacles, boundaries, and free-space layout. The fused BEV representation combines these complementary cues in a common top-down space.

The fused BEV feature is high-dimensional and difficult to interpret from a single pooled map. We therefore inspect selected channels in Figure 13. Different channels exhibit different activation patterns: some are more geometry-dominant, while others reflect broader semantic or contextual responses. This heterogeneity supports the use of query-conditioned aggregation in BEVTraj, because the model must select the regions and feature channels that are relevant to the target agent and its candidate future motions.

### 4.4.3. *Dense scene-level motion modeling*

Beyond the designated target agent, BEVTraj uses dense future prediction to encourage the scene context encoder to capture surrounding-agent motion tendencies, following the practice of MTR (Shi *et al.*, 2022). During scene-context encoding, the model predicts dense multi-agent futures over the BEV representation, providing auxiliary interaction cues that are later used for target-agent forecasting. As shown in Figure 14, the dense future prediction provides an intermediate view of how surrounding agents are expected to move. This auxiliary prediction is not the final output of the model, but it helps the scene context encoder represent interaction-related information before the target-agent decoder produces multimodal trajectories.

### 4.4.4. *Layer-wise evolution of deformable attention geometry*

To examine how spatial evidence evolves during trajectory refinement, we visualize the layer-wise reference points and deformable sampling locations of ITR in Figure 15. As the trajectory hypothesis is progressively updated, the sampling locations shift accordingly, allowing each refinement layer to retrieve BEV features that remain spatially aligned with the latest predicted path. At the same time, different future waypoints query different local regions along the trajectory, enabling long-horizon refinement to use spatial evidence specific to each portion of the predicted motion rather than relying on a fixed or single spatial reference.

## 4.5. Ablation study

To better understand how BEVTraj achieves its overall performance, we conduct a comprehensive series of ablation studies under a consistent experimental protocol. Rather than evaluating the framework only by adding or removing entire modules, the analysis examines the mechanisms underlying its major components and compares the adopted designs with plausible alternative implementations. This allows us to determine whether the proposed mechanisms behave as intended, attribute the observed performance to specific design decisions, and clarify the empirical rationale for the final model configuration. Unless otherwise stated, all experiments follow the same training and evaluation protocol as the main comparison.

### 4.5.1. *Analysis of sparse goal candidate proposal*

Table 6 examines the main design choices of SGCP and shows that the proposed AGGP and OMAC contribute to improved multimodal prediction. AGGP consistently provides strong performance by avoiding the limitations of both direct regression and discrete goal-probability prediction. Direct regression requires a single network to jointly interpret semantic BEV features and geometric goal locations, whereas categorical prediction does not explicitly model geometric relationships among anchors and discards information from neighboring plausible candidates when selecting a single goal. AGGP instead uses learned anchor compatibility to aggregate multiple anchor positions into a continuous goal, combining semantic relevance with explicit spatial structure.

OMAC further improves the top-ranked metrics by encouraging different mode queries to specialize in distinct spatial regions, thereby promoting mode diversity and discriminability. However, without OMAC, unrestricted modes were observed to concentrate on a few dominant motion directions, which improves some best-of-10 metrics but reduces the representation of alternative feasible futures. Such mode concentration can be undesirable for downstream autonomous-driving decisions. Additionally, Centroid-based positive-mode assignment substantially degrades multimodal performance because fixed cluster centroids provide only coarse coverage of the endpoint space, such that diverse future motions may be assigned to modes whose representative centroids are spatially distant from their actual endpoints (Shi *et al.*, 2022).

### 4.5.2. *Trajectory-aware mode scoring*

Table 7 shows that trajectory-aware mode scoring improves mode ranking over direct scoring from decoder embeddings, leading to substantial gains particularly in $\text{minADE}_5$, $\text{minFDE}_1$, and $\text{MR}_5$. This design is particularly well matched to BEVTraj's trajectory-local BEV retrieval mechanism, which gathers spatial information around the current predicted trajectory rather than repeatedly accessing a globally structured scene representation. As ITR updates the trajectory, the relevant spatial evidence changes accordingly; re-encoding the refined trajectory and re-sampling BEV features along its updated waypoints therefore allows the mode confidence to reflect the latest trajectory hypothesis. The benefit is already evident with a single ITR layer, while increasing the refinement depth itself does not produce monotonic improvements across the evaluated metrics.

### 4.5.3. *Temporal consistency in trajectory refinement*

Figure 16 qualitatively illustrates the effect of the proposed temporal-consistency mechanisms on trajectory refinement.

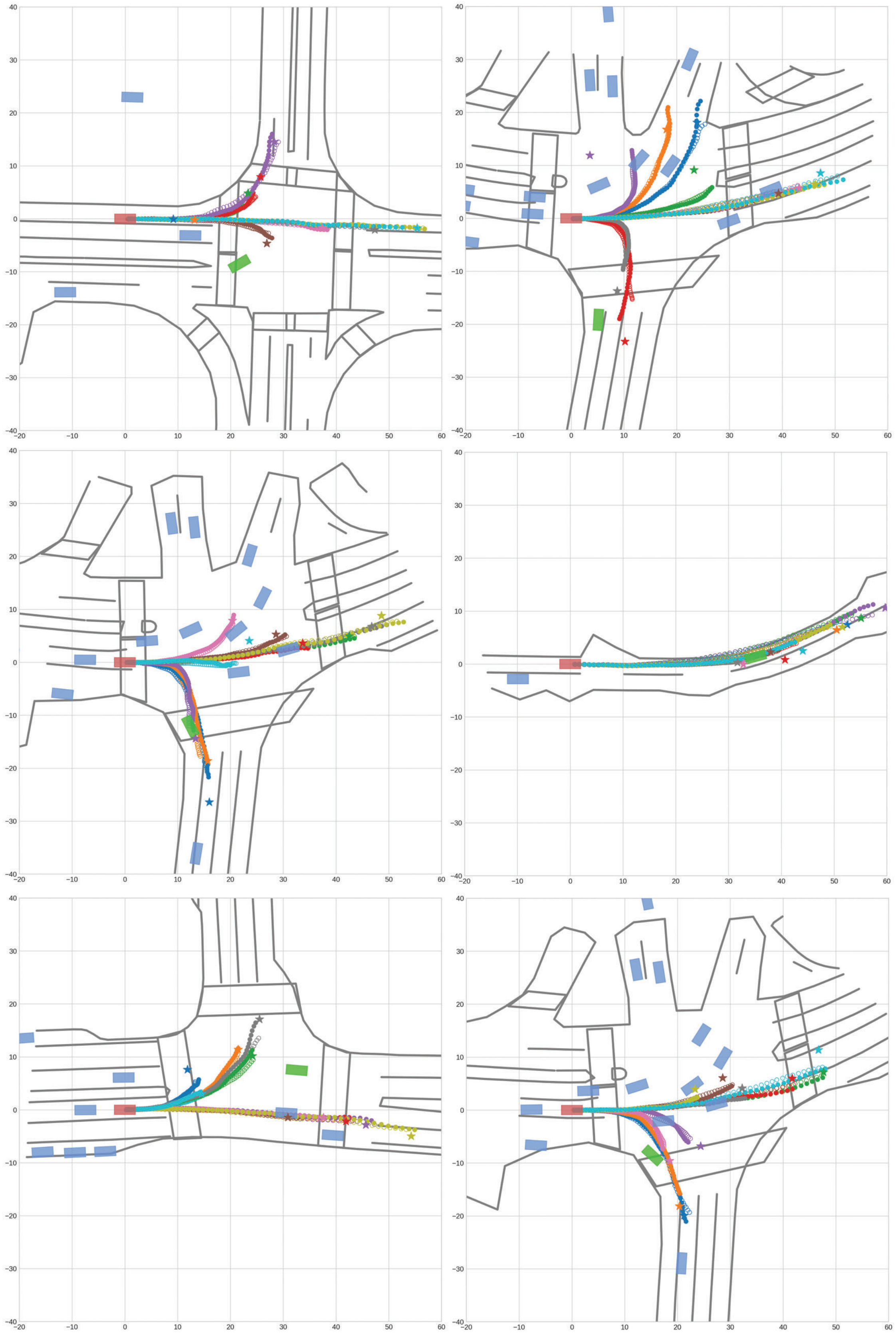

**Figure 11:** Visualization of SGCP-proposed goals and the corresponding decoded trajectories across representative scenes. Star markers denote the goals predicted by SGCP, while hollow and filled circles represent the initial trajectories produced by ITP and the final trajectories after iterative refinement, respectively. Different colors indicate different motion modes, with each goal–trajectory pair sharing the same color. The red and green rectangles denote the target agent and the ego vehicle serving as the reference for sensor observations, respectively.

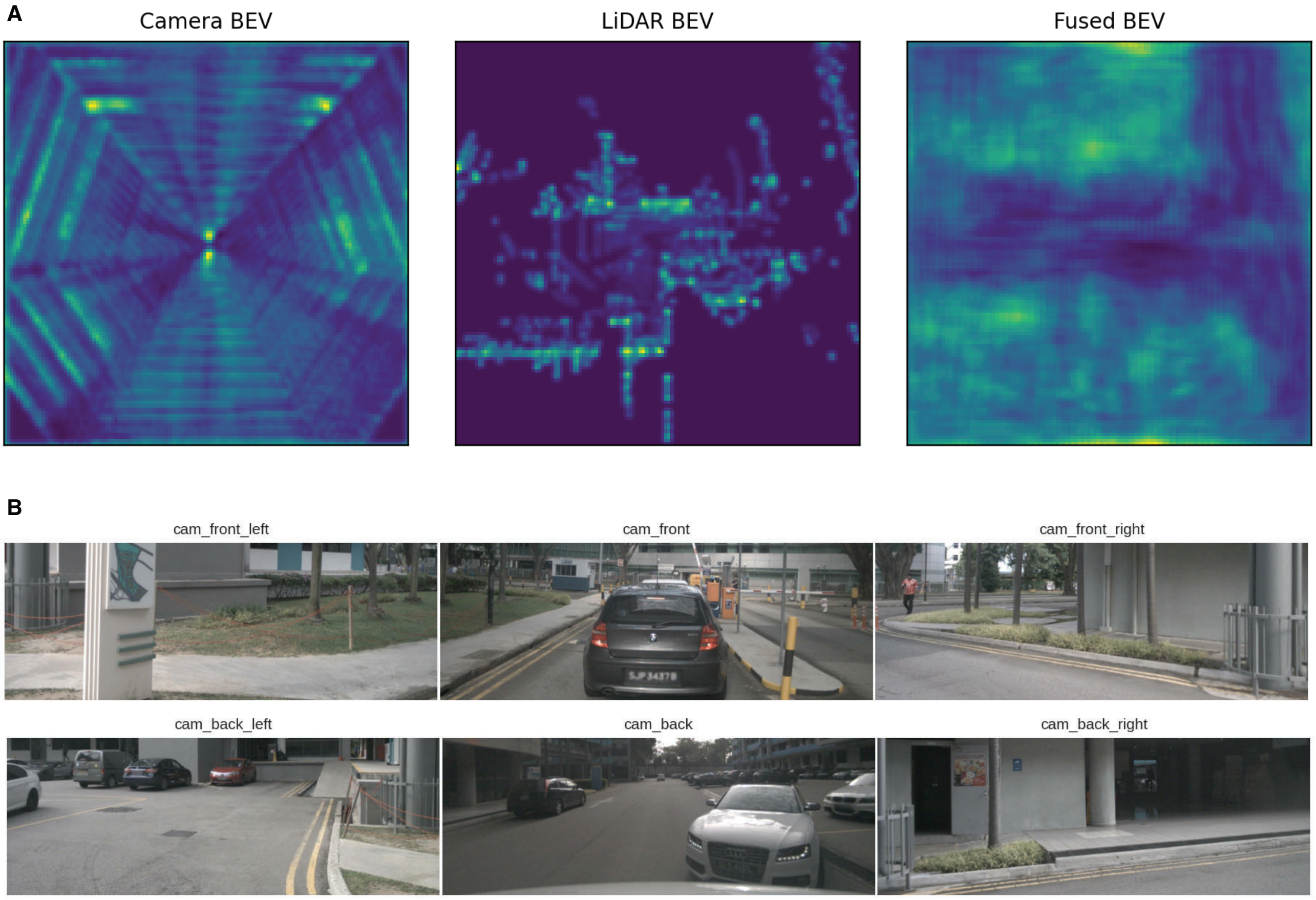

**Figure 12:** Qualitative visualization of the BEV representation prior to trajectory reasoning. (A) Fused BEV feature representation constructed from multi-view cameras and LiDAR, where the right side corresponds to the ego vehicle's forward direction. (B) Corresponding multi-view camera images used jointly with LiDAR to build the fused BEV representation.

Without these mechanisms, independently predicted waypoint-wise residual updates can accumulate into locally irregular trajectory shapes, whereas Gaussian Refinement Smoothing together with trajectory smoothness supervision produces more temporally coherent trajectories while preserving distinct multimodal motion hypotheses. Table 8 further evaluates the two mechanisms quantitatively. Their joint use yields modest improvements across most of the reported metrics. These gains are notable because the mechanisms are introduced primarily to regularize temporal consistency rather than to directly optimize the reported displacement metrics, suggesting that suppressing irregular local refinement updates is also beneficial to prediction accuracy.

### 4.5.4. *Additional design choices*

Table 9 evaluates several additional design choices in BEVTraj. The number of encoder-side BDA queries, which grows proportionally with the number of agents around which local anchor templates are placed, has a comparatively limited effect on prediction performance. This behavior can be attributed to the target agent being included in every configuration. Even with fewer queries, the anchors placed around the target agent cover regions associated with its plausible future motion, allowing BDA to retrieve the minimum spatial context required for reliable prediction. Adding anchors around more neighboring agents provides broader scene context, but the results suggest that this additional coverage is less influential than adequately representing the target agent's reachable region.

The remaining ablations further demonstrate the importance of the motion and geometric priors used at different prediction stages. Target-motion conditioning provides SGCP with a minimal spatial prior derived from the target agent's observed history, helping its predicted goals remain consistent with the agent's recent motion. Predefined trajectories are useful before an initial trajectory has been predicted because their waypoints provide path-wise reference locations for deformable attention; this allows the decoder to inspect spatial evidence along a plausible route toward the goal rather than only around the goal region. The Pre-Encoder is also beneficial because subsequent scene-context attention operates on agent features whose temporal dimension has already been compressed. Applying temporal attention and timestep-wise social attention beforehand preserves the evolution of agent motion and interactions over the observed history, providing a stronger basis for modeling their influence across the future prediction horizon. Finally, content–position decoupling in ITR is particularly relevant to the geometric nature of trajectory prediction. Instead of adding positional encodings element-wise to content features before attention, as in conventional deformable attention, the adopted design computes the attention scores using separate content-based and position-based dot products. The resulting improvement supports explicitly modeling the spatial correspondence between trajectory waypoints and sampled BEV features during trajectory refinement.

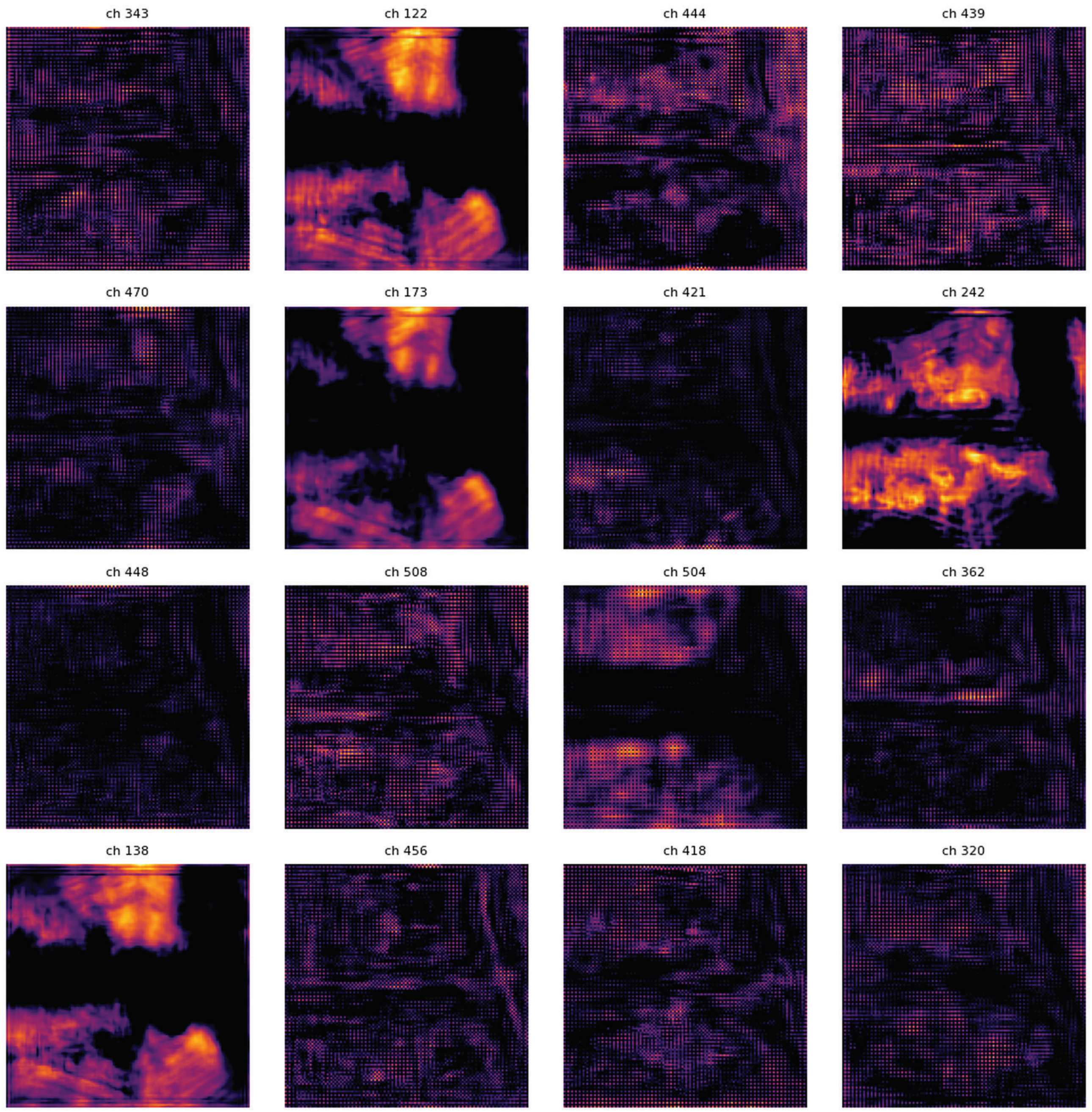

**Figure 13:** Per-channel visualization of the fused BEV feature. Different channels exhibit distinct activation patterns, indicating that the fused representation contains heterogeneous semantic and geometric cues.

## 4.6. Computational efficiency and scalability

This subsection examines the computational efficiency and scalability of BEVTraj in conjunction with the predictive effectiveness of its key architectural designs. As an overall reference for the computational footprint of BEVTraj, the complete model requires 188.007 ms for end-to-end inference, 503.027 G FLOPs, and approximately 2.23 GiB of peak inference memory, while the prediction pipeline excluding the BEVFusion sensor encoder requires 65.424 ms. Since the main methodological components of BEVTraj are concentrated in the decoder, the detailed computational comparisons below focus primarily on decoder-side inference cost, while full-model training memory is also reported to characterize the corresponding training cost.

BEVTraj adopts deformable attention throughout the model to selectively retrieve relevant spatial evidence from dense BEV features without the cost of exhaustively attending to the full spatial grid. To verify that this architectural choice is both computationally efficient and effective for trajectory prediction, we conduct two complementary comparisons. Table 10 first compares alternative BEV feature-retrieval mechanisms under an approximately equal computational budget, isolating the effectiveness of the attention design from differences in computational cost. For budget matching, the global-attention baseline reduces the number of BEV tokens through adaptive max pooling and adjusts the attention-head configuration while preserving global attention over all pooled tokens and keeping the overall key/value dimension unchanged; all other model settings remain fixed.

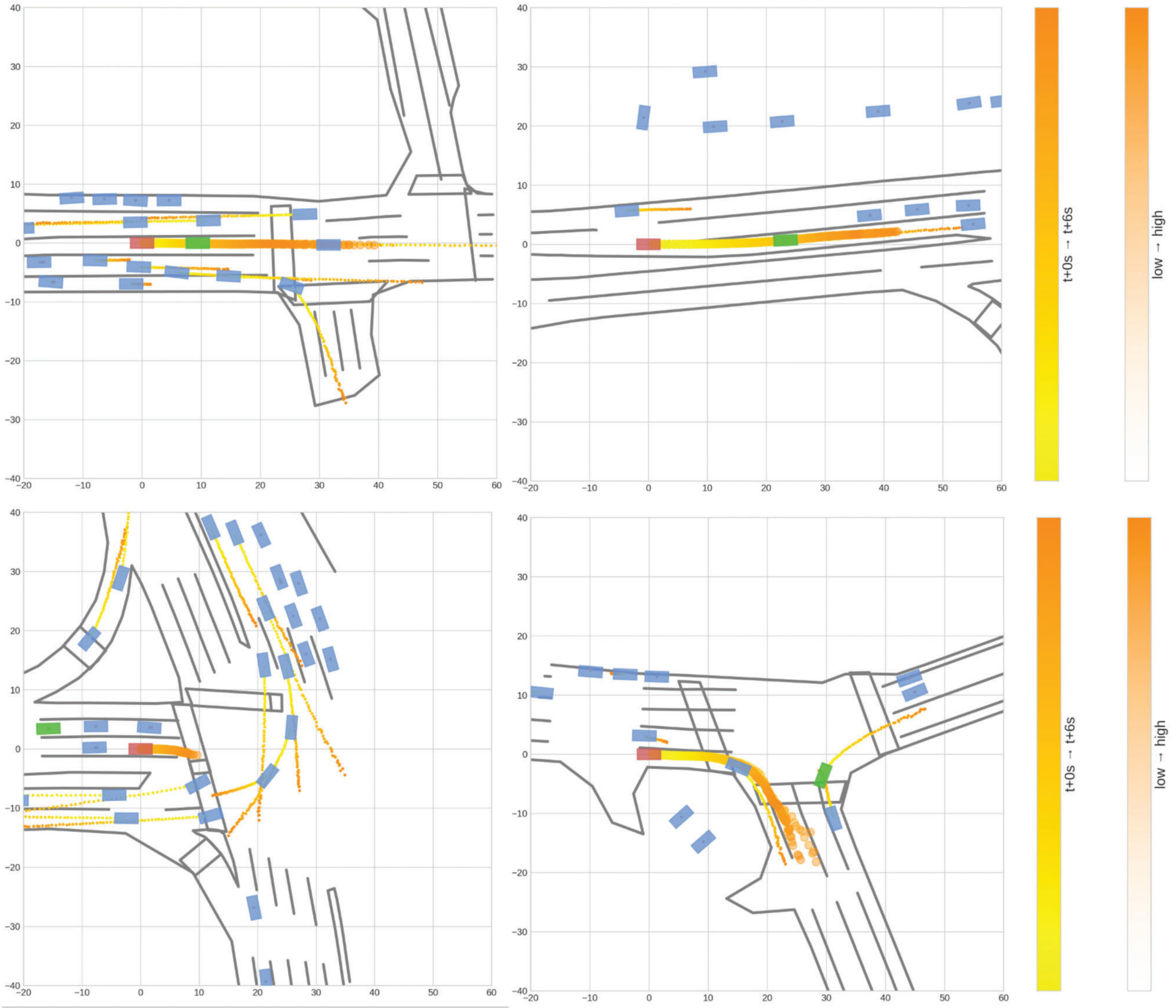

**Figure 14:** Visualization of dense future prediction for surrounding agents. This intermediate output provides additional interaction-related context for target-agent motion forecasting.

Deformable attention achieves the best overall performance, suggesting that it retrieves prediction-relevant BEV evidence more effectively than the vanilla alternatives. This result is consistent with its trajectory-centered feature-retrieval mechanism: predicted trajectory waypoints provide geometrically aligned reference centers, while learned sampling offsets allow the model to gather scene-adaptive evidence around and beyond them. In contrast, global vanilla attention lacks such trajectory-aligned spatial guidance, whereas local vanilla attention is constrained to a fixed region and therefore cannot adapt its retrieval pattern to each predicted trajectory. Table 11 then evaluates the computational cost of a full-grid global vanilla-attention implementation without constraining it to BEVTraj's computational budget. Explicitly attending to all BEV locations increases decoder latency, FLOPs, and peak inference memory; it also requires substantially more memory for end-to-end training, demonstrating the practical efficiency of BEV feature retrieval using deformable attention. Together, the two comparisons show that deformable attention retains its overall predictive advantage when computational capacity is controlled while reducing computational cost in its native configuration.

Beyond the efficiency of BEV feature retrieval, we further examine the sensitivity of BEVTraj to the number of decoded trajectory modes, $K$. Since each mode corresponds to a sparse goal hypothesis in SGCP and a trajectory to be decoded, varying $K$ also directly scales the prediction workload and provides a task-specific view of model scalability. SGCP is designed to avoid treating every spatial goal candidate as a separate trajectory branch. By using a broad shared anchor set for spatial goal exploration while decoding only a compact set of output modes, SGCP aims to preserve multimodal spatial coverage without incurring the computational and memory costs of decoding a large number of trajectories.

We therefore vary the number of output modes and examine both prediction performance and computational cost. To keep the degree of overlap between OMAC clusters approximately comparable across settings, we reduce the number of anchors assigned to each cluster as the number of modes increases, thereby narrowing the spatial region covered by each mode. As shown in Tables 12 and 13, increasing the number of decoded trajectories results in only modest changes in prediction accuracy while substantially increasing decoder latency, FLOPs, and memory con-sumption. The $K = 10$ configuration therefore provides a favorable balance between predictive performance and computational efficiency while directly producing the required number of output trajectories without additional mode-reduction post-processing such as NMS. These results support the central motivation of SGCP: broad spatial coverage does not require a correspondingly large set of explicitly decoded trajectories.

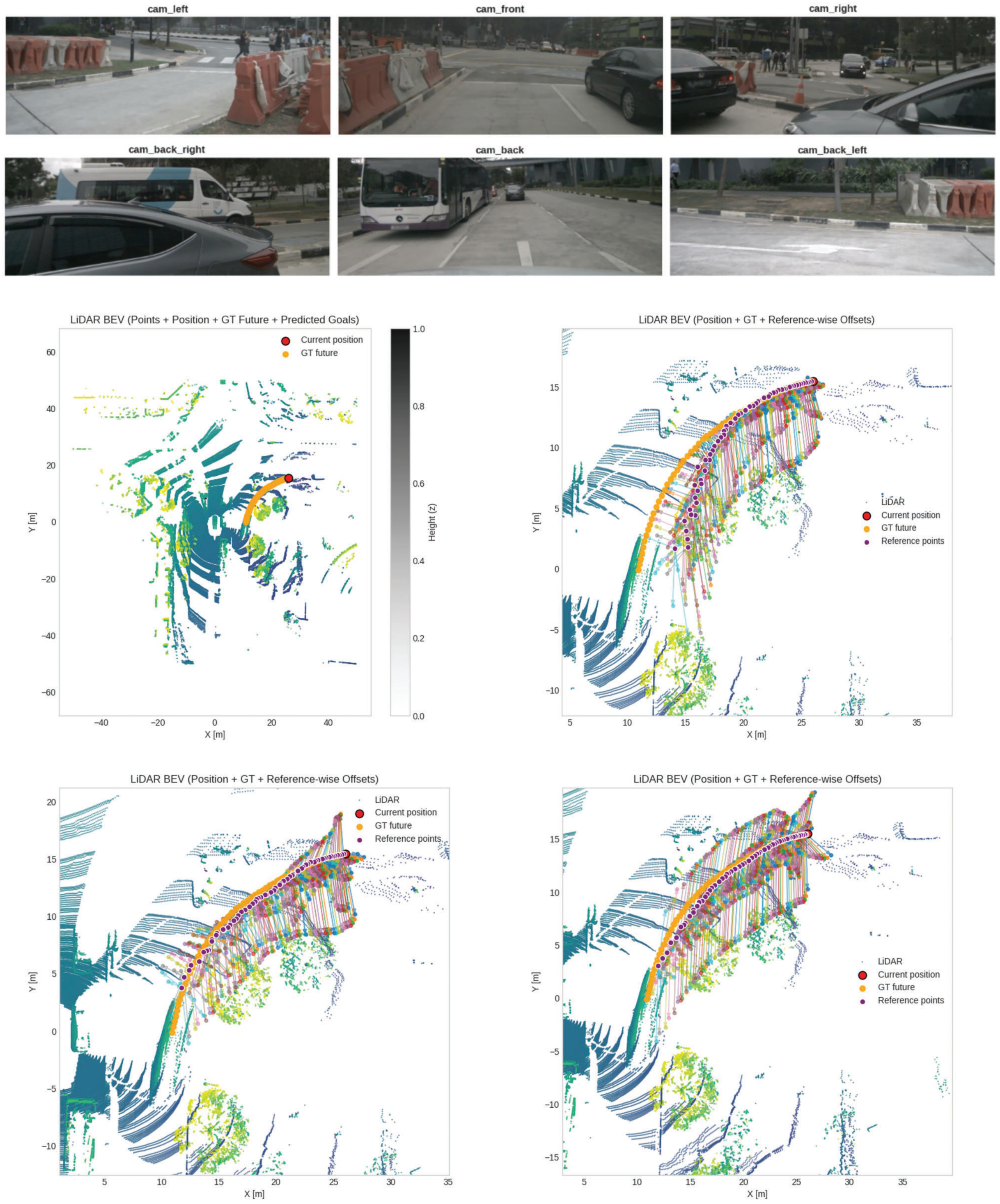

**Figure 15:** Visualization of the layer-wise evolution of predicted waypoints and deformable-attention sampling locations in the Iterative Trajectory Refinement (ITR) module. The top row shows the six surrounding camera views. The first BEV panel provides a scene-level LiDAR overview with the current target position, ground-truth future trajectory, and predicted motion hypotheses. The remaining three panels correspond to successive ITR layers, showing the predicted waypoints as deformable-attention reference points with their associated sampling locations and the ground-truth trajectory.

**Table 6:** Ablation study of the SGCP design on the *nuScenes* validation set. OMAC denotes overlapping mode-wise anchor clusters. *Reg.*, *Prob.*, and *AGGP* denote direct goal regression, goal-probability prediction, and attention-guided goal prediction, respectively. *Pred.* and *Cent.* denote positive-mode assignment using the predicted goal and cluster centroid. Without SGCP, learnable mode queries are paired with fixed trajectory prototypes. Bold entries indicate the configuration used in the original BEVTraj model.

| OMAC | Goal | Assign. | $\text{minADE}_5\downarrow$ | $\text{minADE}_{10}\downarrow$ | $\text{minFDE}_1\downarrow$ | $\text{MR}_5\downarrow$ | $\text{MR}_{10}\downarrow$ |
|---|---|---|---|---|---|---|---|
| – | w/o SGCP | – | 1.2974 | 1.2513 | 6.8856 | 0.7626 | 0.7559 |
| X | Reg. | Pred. | 1.1692 | 0.9019 | 7.6872 | 0.5294 | 0.4408 |
| O | Reg. | Pred. | 1.1383 | 0.8834 | 7.1521 | 0.5348 | 0.4416 |
| O | Reg. | Cent. | 1.2519 | 1.1444 | 6.5782 | 0.6759 | 0.6545 |
| X | Prob. | Pred. | 1.5301 | 1.1046 | 7.6828 | 0.6107 | 0.4750 |
| O | Prob. | Pred. | 1.3710 | 1.0571 | 7.1573 | 0.5821 | 0.5085 |
| X | AGGP | Pred. | 1.1826 | 0.8359 | 7.2437 | 0.4959 | 0.3453 |
| **O** | **AGGP** | **Pred.** | **1.0543** | **0.8708** | **6.7496** | **0.4745** | **0.4207** |

**Table 7:** Ablation study of mode scoring and ITR depth on the *nuScenes* validation set. *Dec. Embed.* denotes direct mode scoring from decoder embeddings, whereas *Refined Traj.* uses the refined trajectories together with re-sampled BEV context for mode scoring. ITR depth denotes the number of iterative trajectory refinement layers following ITP. Bold entries indicate the configuration used in the original BEVTraj model.

| Scoring | ITR Depth | $\text{minADE}_5\downarrow$ | $\text{minADE}_{10}\downarrow$ | $\text{minFDE}_1\downarrow$ | $\text{MR}_5\downarrow$ | $\text{MR}_{10}\downarrow$ |
|---|---|---|---|---|---|---|
| Dec. Embed. | 1 | 1.2079 | 0.9072 | 7.9339 | 0.5034 | 0.4282 |
| Dec. Embed. | 3 | 1.1683 | 0.8690 | 7.7414 | 0.4889 | 0.4053 |
| Dec. Embed. | 5 | 1.1774 | 0.8803 | 7.9286 | 0.4892 | 0.4091 |
| Refined Traj. | 1 | **1.0452** | 0.9023 | 6.8162 | 0.4918 | 0.4503 |
| Refined Traj. | 3 | 1.0543 | 0.8708 | **6.7496** | 0.4745 | 0.4207 |
| Refined Traj. | 5 | 1.0549 | **0.8680** | 6.7983 | **0.4737** | **0.4045** |

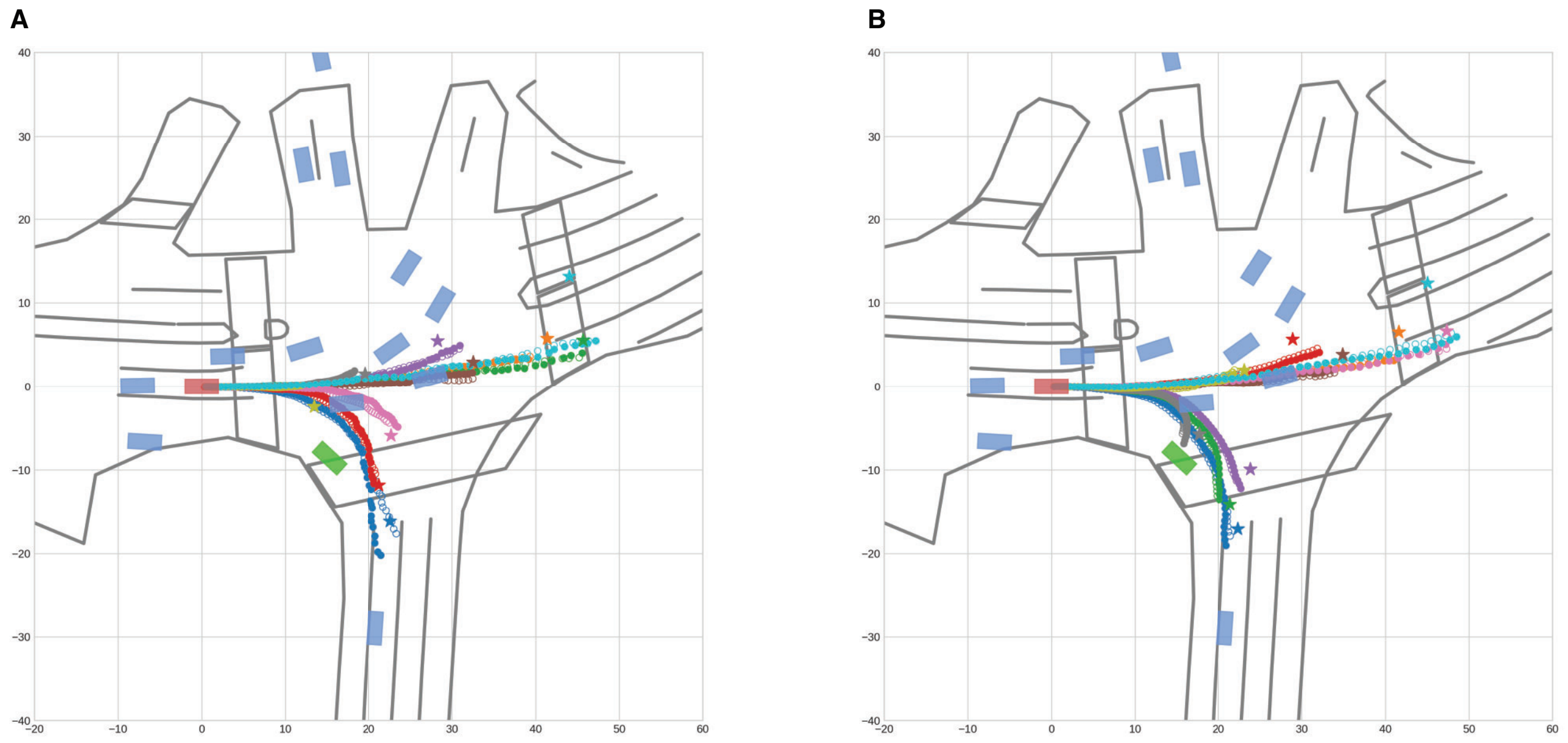

**Figure 16:** Qualitative effect of the proposed temporal-consistency mechanisms. (A) Without temporal-consistency mechanisms.(B) With the proposed Gaussian refinement smoothing and trajectory smoothness supervision. The proposed mechanisms suppress locally irregular waypoint updates while preserving diverse multimodal trajectory hypotheses.

## 5. Discussion

The preceding sections have established the design and empirical behavior of BEVTraj through quantitative comparisons, qualitative analyses, and detailed ablation studies. However, these evaluations alone do not fully capture the broader implications of formulating trajectory prediction around sensor-derived BEV rep-

**Table 8:** Ablation study of temporal consistency modeling in trajectory refinement on the *nuScenes* validation set. *Gaussian* denotes Gaussian smoothing of the predicted refinement offsets, and *Smooth. Loss* denotes the trajectory smoothness loss based on acceleration, jerk, and motion-direction consistency. Bold values indicate the best (lowest) performance for each metric.

| Gaussian | Smooth. Loss | $\text{minADE}_5\downarrow$ | $\text{minADE}_{10}\downarrow$ | $\text{minFDE}_1\downarrow$ | $\text{MR}_5\downarrow$ | $\text{MR}_{10}\downarrow$ |
|---|---|---|---|---|---|---|
| X | X | **1.0426** | 0.8785 | 6.8396 | 0.4828 | 0.4333 |
| X | O | 1.0554 | 0.8807 | 6.7686 | 0.4822 | 0.4223 |
| O | X | 1.0851 | 0.8819 | 7.1255 | 0.4954 | 0.4344 |
| O | O | 1.0543 | **0.8708** | **6.7496** | **0.4745** | **0.4207** |

**Table 9:** Ablation study of additional design choices on the nuScenes validation set. Bold values indicate the best (lowest) performance for each metric within each ablation group.

| Configuration | $\text{minADE}_5\downarrow$ | $\text{minADE}_{10}\downarrow$ | $\text{minFDE}_1\downarrow$ | $\text{MR}_5\downarrow$ | $\text{MR}_{10}\downarrow$ |
|---|---|---|---|---|---|
| *Number of BDA encoder queries* | | | | | |
| 128 queries ($N_o = 2$) | 1.0812 | **0.8656** | 7.1047 | 0.4773 | **0.4042** |
| 256 queries ($N_o = 4$) | 1.0573 | 0.8761 | 6.8494 | **0.4668** | 0.4068 |
| 512 queries ($N_o = 8$, BEVTraj) | 1.0543 | 0.8708 | 6.7496 | 0.4745 | 0.4207 |
| 1024 queries ($N_o = 16$) | **1.0532** | 0.8825 | **6.6841** | 0.4779 | 0.4186 |
| *Other component choices* | | | | | |
| w/o target-motion conditioning in SGCP | 1.1150 | 0.9409 | 6.7728 | 0.4884 | 0.4395 |
| w/o predefined trajectories | 1.0663 | 0.8774 | 7.1419 | 0.4771 | **0.4186** |
| w/o pre-encoder | 1.0820 | 0.8957 | 6.8737 | 0.4822 | 0.4248 |
| w/o content–position decoupling in ITR | 1.0996 | 0.9088 | **6.6919** | 0.4977 | 0.4439 |
| Full BEVTraj | **1.0543** | **0.8708** | 6.7496 | **0.4745** | 0.4207 |

**Table 10:** Comparison of BEV feature-retrieval mechanisms under an approximately matched computational budget on the *nuScenes* validation set. Bold values indicate the best (lowest) performance for each metric.

| Attention mechanism | $\text{minADE}_5\downarrow$ | $\text{minADE}_{10}\downarrow$ | $\text{minFDE}_1\downarrow$ | $\text{MR}_5\downarrow$ | $\text{MR}_{10}\downarrow$ |
|---|---|---|---|---|---|
| Global vanilla attention | 1.1189 | 0.9580 | **6.6946** | 0.4889 | 0.4480 |
| Local vanilla attention (fixed local region) | 1.1446 | 0.8998 | 7.3704 | 0.5090 | 0.4243 |
| Deformable attention (BEVTraj) | **1.0543** | **0.8708** | 6.7496 | **0.4745** | **0.4207** |

**Table 11:** Direct computational comparison between deformable attention and global vanilla attention over the full BEV grid. All measurements are obtained on a single RTX 4090 with a batch size of 1.

| Model | | Decoder inference | | Full BEVTraj training |
|---|---|---|---|---|
| | Latency (ms) | FLOPs (G) | Peak memory (MiB) | Peak memory (MiB) |
| Deformable attention (BEVTraj) | 46.289 | 111.845 | 660.956 | 7378.93 |
| Global vanilla attention | 57.467 | 243.174 | 3356.029 | 19,114.92 |

**Table 12:** Effect of the number of decoded trajectory modes on the *nuScenes* validation set. For configurations with more than 10 modes, NMS is applied to retain 10 trajectories for evaluation under the same prediction protocol.

| Decoded modes | $\text{minADE}_5\downarrow$ | $\text{minADE}_{10}\downarrow$ | $\text{minFDE}_1\downarrow$ | $\text{MR}_5\downarrow$ | $\text{MR}_{10}\downarrow$ |
|---|---|---|---|---|---|
| 10 | 1.0543 | 0.8708 | 6.7496 | 0.4745 | 0.4207 |
| 32 | 1.0978 | 0.8542 | 6.7323 | 0.4755 | 0.3988 |
| 64 | 1.0975 | 0.8264 | 6.8091 | 0.4840 | 0.3957 |

**Table 13:** Computational scalability with respect to the number of decoded trajectory modes. All measurements are obtained on a single RTX 4090 with a batch size of 1.

| Modes | | Decoder inference | | Full model training |
|---|---|---|---|---|
| | Latency (ms) | FLOPs (G) | Peak memory (MiB) | Peak memory (MiB) |
| 10 | 46.289 | 111.845 | 660.956 | 7378.93 |
| 32 | 52.958 | 201.618 | 773.232 | 11 890.55 |
| 64 | 81.687 | 351.630 | 2012.783 | 21 311.31 |

**Table 14:** Ego-vehicle trajectory forecasting results on the *nuScenes* validation set under the same spatial-coverage constraint.

| Method | $\mathrm{minADE}_5$ ↓ | $\mathrm{minADE}_{10}$ ↓ | $\mathrm{minFDE}_1$ ↓ | $\mathrm{MR}_5$ ↓ | $\mathrm{MR}_{10}$ ↓ |
|---|---|---|---|---|---|
| Autobot | 1.4037 | 0.8004 | 6.7718 | 0.4626 | 0.3195 |
| MTR | 0.9589 | 0.8556 | 5.8999 | 0.4816 | 0.4517 |
| Wayformer | 1.0944 | 0.7184 | 6.2869 | 0.4248 | 0.2894 |
| DeMo | 1.0685 | 0.6945 | 5.0480 | 0.3907 | 0.2410 |
| BEVTraj (ours) | 0.9852 | 0.7087 | 5.7196 | 0.4025 | 0.3105 |

resentations. Further consideration is needed to understand the scope of the proposed framework, its potential beyond the settings directly examined in the experiments, and the challenges associated with its broader application. This section complements the preceding empirical analysis by providing a broader perspective on the interpretation, applicability, and future development of BEVTraj.

## 5.1. Interpreting the map-free formulation

BEVTraj is HD-map-free, but it is not structure-free. Instead of using manually curated map priors such as lane topology and road boundaries, it uses sensor-derived BEV features as an implicit spatial representation. Accordingly, the primary objective of this work is not to argue that pre-built map information is unnecessary for motion forecasting, but to investigate whether reliable agent-level forecasting can be achieved from online sensor-derived spatial context when such map information is unavailable, outdated, or difficult to maintain. In this sense, BEVTraj replaces explicit HD-map priors with online spatial cues inferred from sensor observations, rather than attempting to remove environmental structure from the prediction problem.

This distinction is important for interpreting the experimental results. Sensor-derived BEV features provide rich local geometric and semantic cues, including obstacles, free space, road boundaries, and road-surface patterns, but their spatial coverage is fundamentally constrained by the observable sensor range. Conversely, pre-built maps can provide broader and more persistent structural priors, particularly for long-range road topology and traffic-rule information. The two sources of information should therefore be viewed as complementary rather than mutually exclusive. In particular, an interesting extension of BEVTraj would be to combine its locally detailed sensor-derived BEV representation with coarser but broader-coverage Standard-Definition or navigation-map priors, which have been shown to provide useful road-level context for motion prediction (Dong *et al.*, 2025; Schmidt *et al.*, 2023). Such a hybrid formulation could preserve the ability to operate without lane-level HD-map infrastructure while supplementing the limited spatial coverage of onboard sensing with longer-range topological context.

## 5.2. Ego-vehicle trajectory prediction and planning relevance

In addition to surrounding-agent prediction, we evaluated ego-vehicle trajectory forecasting under the same spatial-coverage constraint and found that BEVTraj achieves competitive performance, as shown in Table 14. These results suggest that sensor-derived BEV features can also provide useful context for ego-motion forecasting, where onboard observations are directly aligned with the ego vehicle's motion and immediate surroundings.

The proposed framework is not a planning method, and the experiments do not evaluate closed-loop planning performance. Nevertheless, explicit multimodal trajectory outputs can be used as inputs to downstream planning modules, for example by converting predicted trajectories into occupancy or risk maps, checking candidate plans against predicted agent motions, or constructing a small set of possible future scenarios. A full integration with a planner is beyond the scope of this paper, but the BEV-based formulation and explicit trajectory outputs make such integration feasible.

Beyond planner-integrated applications, another potential direction is human-centered intelligent-vehicle interaction. Explicit multimodal trajectory predictions could potentially be used to communicate predicted traffic developments and associated uncertainty to human drivers or supervisors. Previous work has shown the importance of human–machine interface design in avoiding mode confusion in automated vehicles (Eom & Lee, 2022). In addition, robust gaze estimation using attention-based models has been investigated under challenging head-pose and gaze variations (Zhong & Lee, 2025), suggesting the possibility of combining external scene forecasting with driver-attention or intention cues in future human-centered driving systems.

## 5.3. Extension to multi-agent scene-level analysis

BEVTraj is primarily designed as a target-agent trajectory predictor. However, because it outputs multiple trajectory hypotheses for each agent, the predictions can also be combined into joint future hypotheses for multiple agents. This can be useful for analyzing interaction consistency and collision-related behavior at the scene level. In this work, this procedure is treated as a lightweight post-processing extension rather than a core component of the BEVTraj architecture. The detailed formulation, including the joint-world assignment cost and supplementary results, is provided in A.

## 5.4. Limitations

Evaluating sensor-based trajectory prediction presents an inherent challenge because prediction performance depends not only on the prediction model itself but also on the availability and quality of the observed scene information. This issue becomes particularly important in our comparison with pre-built-HD-map-based methods, which provide the closest agent-centric prediction counterparts given the lack of directly comparable agent-centric approaches operating purely on sensor observations. Nevertheless, the two paradigms cannot be placed under completely equivalent input conditions. Sensor-derived representations are directly affected by occlusion, increasing sensing distance, and adverse environmental conditions such as nighttime or severe weather, whereas pre-built HD maps are largely unaffected by such sensing degradation. Ideally, these factors would be explicitly controlled when constructing the evaluation set so that differences in trajectory prediction capability can be isolated from differences in input observability. However, jointly accounting for the many scene- and sensor-dependent factors that determine observation quality is difficult in practice. Consequently, the current evaluation may still conflate prediction errors caused by limited or degraded sensor evidence with those arising from the trajectory prediction model itself. Developing evaluation protocols that more precisely separate these effects would therefore allow the prediction capability of BEVTraj and other sensor-based approaches to be assessed more independently of upstream sensing quality.

Beyond improved evaluation protocols, future work may also consider robustness to domain shifts and uncertainties in agent association. Domain-adaptation techniques have been explored for transferring vehicle representations from synthetic to real visual domains (Lee *et al.*, 2020), while attention-based vehicle re-identification has been used to improve discriminative vehicle representations under appearance and viewpoint variations (Lee *et al.*, 2023). Integrating such domain-adaptive and identity-aware representations with sensor-derived BEV forecasting may help maintain more reliable spatial context and agent histories across changing environments, sensing conditions, and partial occlusions.

The proposed prediction architecture also retains several structural limitations. OMAC enables a compact number of modes to maintain a useful balance between scene adaptivity and multimodal diversity by allowing each mode to reason over an overlapping local anchor region. Nevertheless, the spatial support assigned to each mode is still predefined. If a mode's anchor cluster lies largely outside the drivable region of a particular scene, the corresponding mode cannot dynamically relocate its candidate region to a more plausible part of the scene, potentially leaving part of the available prediction capacity ineffective. In addition, although AGGP and waypoint-wise residual refinement reduce the burden of directly inferring future geometry, BEVTraj still represents agent motion in a vectorized form while scene context is encoded primarily through sensor-derived BEV representations. The model must therefore reason jointly over representations with substantially different structures and semantics during scene-context encoding and trajectory decoding. Developing representations or interaction mechanisms that more naturally bridge vectorized motion states and dense sensor-derived scene features, together with more scene-adaptive allocation of goal-search regions, constitutes an important direction for future work.

## 6. Conclusion

This paper presented BEVTraj, a map-free trajectory prediction framework that directly exploits sensor-derived BEV representations for agent-centric multimodal forecasting without relying on pre-built HD maps. The central challenge of this formulation is to extract motion-relevant spatial information from dense and implicitly structured BEV features and convert it into compact, scene-adaptive future hypotheses. BEVTraj addresses this challenge by using deformable attention to selectively retrieve prediction-relevant spatial evidence from dense BEV features for each target agent. To represent multimodal futures efficiently, SGCP decouples broad spatial goal coverage from the number of explicitly decoded trajectory modes, allowing a compact set of modes to infer scene-adaptive goals from a larger shared set of spatial anchors. The resulting goal hypotheses are transformed into full trajectories and progressively refined using trajectory-aligned BEV evidence, with trajectory-aware mode scoring and temporal-consistency modeling supporting reliable mode evaluation and temporally coherent motion throughout refinement.

Extensive experiments on *nuScenes* and Argoverse 2 Sensor provide evidence for the effectiveness of this formulation. Under a controlled local-input protocol aligned with the spatial coverage available to sensor-derived BEV perception, BEVTraj achieves competitive multimodal trajectory prediction performance against representative HD-map-based approaches. The accompanying map-range analysis further characterizes the effect of spatial context available to the map-based baselines, while the complementary end-to-end occupancy-style evaluation shows that the predicted agent trajectories also preserve coherent scene-level spatial structure. Qualitative analyses illustrate how BEVTraj adapts its spatial reasoning to different motion hypotheses and evolving trajectories. Furthermore, mechanism-level ablations validate the major decoder and goal-prediction designs, and the computational analyses show that the use of deformable attention and compact multimodal decoding provides an effective balance between prediction performance, spatial coverage, and computational cost.

Taken together, these results suggest that sensor-derived BEV representations can serve as a practical spatial foundation for agent-level multimodal trajectory prediction, extending their role beyond perception and occupancy modeling. BEVTraj suggests that online geometric and semantic scene evidence can be selectively organized into compact motion hypotheses and refined directly into explicit future trajectories without requiring lane-level HD-map infrastructure. This formulation provides a flexible basis for trajectory prediction in environments where map availability or freshness cannot be assumed, while retaining explicit multimodal outputs that may provide useful inputs for ego-motion forecasting, multi-agent scene analysis, and downstream autonomous-driving modules, subject to further planner-integrated evaluation.

## Conflict of interest

The authors declare no conflict of interest.

## Author contributions

**Minsang Kong:** Conceptualization, Methodology, Software, Data curation, Visualization, Writing–original draft. **Myeongjun Kim:** Conceptualization, Methodology, Software, Data curation, Writing–original draft. **Sang Gu Kang:** Software, Data curation, Validation. **Hejiu Lu:** Methodology, Software, Visualization, Writing–original draft. **Yupeng Zhong:** Investigation, Data curation, Software, Visualization, Writing–original draft. **Sang Hun Lee:** Supervision, Writing–review & editing, Project administration, Funding acquisition.

## Funding

This work was supported by the National Research Foundation of Korea (NRF) grant funded by the Korean government (MSIT) (No. RS-2026-25491225).

## Data availability

The data underlying this article were derived from publicly available third-party datasets. The nuScenes dataset is available from the official nuScenes website at https://www.nuscenes.org/. The Argoverse 2 Sensor dataset is available from the official Argoverse 2 website at https://argoverse.org/av2.html. The derived target-agent sample lists, filtered split indices, and preprocessing meta-data generated from both the nuScenes and Argoverse 2 Sensor datasets in this study will be shared on reasonable request to the corresponding author. No new raw driving sensor data were generated in this study.

The source code is available at https://github.com/Kongminsang/bevtraj.

## References

Akan, A. K., & Güney, F. (2022). Stretchbev: Stretching future instance prediction spatially and temporally. In *European Conference on Computer Vision* (pp. 444–460). Springer.

Biktairov, Y., Stebelev, M., Rudenko, I., Shliazhko, O., & Yangel, B. (2020). Prank: motion prediction based on ranking. *Advances in Neural Information Processing Systems*, **33**, 2553–2563.

Caesar, H., Bankiti, V., Lang, A. H., Vora, S., Liong, V. E., Xu, Q., Krishnan, A., Pan, Y., Baldan, G., & Beijbom, O. (2020). nuscenes: A multimodal dataset for autonomous driving. In *Proceedings of the IEEE/CVF Conference on Computer Vision and Pattern Recognition* (pp. 11621–11631). IEEE.

Casas, S., Gulino, C., Liao, R., & Urtasun, R. (2020). Spagnn: Spatially-aware graph neural networks for relational behavior forecasting from sensor data. In *2020 IEEE International Conference on Robotics and Automation (ICRA)* (pp. 9491–9497). IEEE.

Casas, S., Sadat, A., & Urtasun, R. (2021). Mp3: A unified model to map, perceive, predict and plan. In *Proceedings of the IEEE/CVF Conference on Computer Vision and Pattern Recognition* (pp. 14403–14412). IEEE.

Chai, Y., Sapp, B., Bansal, M., & Anguelov, D. (2019). Multipath: Multiple probabilistic anchor trajectory hypotheses for behavior prediction. *preprint arXiv:1910.05449*, https://doi.org/10.48550/arXiv.1910.05449.

Chen, Y., Liu, J., Zhang, X., Qi, X., & Jia, J. (2023). Largekernel3d: Scaling up kernels in 3d sparse cnns. In *Proceedings of the IEEE/CVF Conference on Computer Vision and Pattern Recognition* (pp. 13488–13498). IEEE.

Cheng, H., Liu, M., Chen, L., Broszio, H., Sester, M., & Yang, M. Y. (2023). Gatraj: A graph-and attention-based multi-agent trajectory prediction model. *ISPRS Journal of Photogrammetry and Remote Sensing*, **205**, 163–175, https://doi.org/10.1016/j.isprsjprs.2023.10.001.

Choi, S., Kim, J., Shin, H., & Choi, J. W. (2024). Mask2map: Vectorized hd map construction using bird's eye view segmentation masks. In *European Conference on Computer Vision* (pp. 19–36). Springer.

Ding, G., Xia, Y., Guan, R., Wu, Q., Huang, T., Ding, W., Sun, J., & Mao, G. (2025). Optipmb: Enhancing 3d multi-object tracking with optimized poisson multi-bernoulli filtering. *preprint arXiv:2503.12968*, https://doi.org/10.48550/arxiv.2503.12968.

Djuric, N., Radosavljevic, V., Cui, H., Nguyen, T., Chou, F.-C., Lin, T.-H., Singh, N., & Schneider, J. (2020). Uncertainty-aware short-term motion prediction of traffic actors for autonomous driving. in *Proceedings of the IEEE/CVF Winter Conference on Applications of Computer Vision*, (pp. 2095–2104). IEEE.

Dong, Z., Ding, R., Li, W., Zhang, P., Tang, G., & Guo, J. (2025). Leveraging sd map to augment HD map-based trajectory prediction. In *2025 IEEE/CVF Conference on Computer Vision and Pattern Recognition (CVPR)* (pp. 17219–17228). IEEE.

Eom, H., & Lee, S. H. (2022). Mode confusion of human–machine interfaces for automated vehicles. *Journal of Computational Design and Engineering*, **9**, 1995–2009, https://doi.org/10.1093/jcde/qwac088.

Fang, S., Wang, Z., Zhong, Y., Ge, J., & Chen, S. (2023). Tbp-former: Learning temporal bird's-eye-view pyramid for joint perception and prediction in vision-centric autonomous driving. in *Proceedings of the IEEE/CVF Conference on Computer Vision and Pattern Recognition*, (pp. 1368–1378). IEEE.

Feng, L., Bahari, M., Amor, K. M. B., Zablocki, É., Cord, M., & Alahi, A. (2024). Unitraj: A unified framework for scalable vehicle trajectory prediction. In *European Conference on Computer Vision* (pp. 106–123). Springer.

Gao, J., Sun, C., Zhao, H., Shen, Y., Anguelov, D., Li, C., & Schmid, C. (2020). Vectornet: Encoding hd maps and agent dynamics from vectorized representation. In *Proceedings of the IEEE/CVF Conference on Computer Vision and Pattern Recognition* (pp. 11525–11533). IEEE.

Girgis, R., Golemo, F., Codevilla, F., Weiss, M., D'Souza, J. A., Kahou, S. E., Heide, F., & Pal, C. (2021). Latent variable sequential set transformers for joint multi-agent motion prediction. *preprint arXiv:2104.00563*, https://doi.org/10.48550/arXiv.2104.00563.

Gu, J., Sun, C., & Zhao, H. (2021). Densetnt: End-to-end trajectory prediction from dense goal sets. in *Proceedings of the IEEE/CVF International Conference on Computer Vision*, (pp. 15303–15312). IEEE.

Hong, J., Sapp, B., & Philbin, J. (2019). Rules of the road: Predicting driving behavior with a convolutional model of semantic interactions. In *Proceedings of the IEEE/CVF Conference on Computer Vision and Pattern Recognition* (pp. 8454–8462). IEEE.

Hou, L., Li, S. E., Yang, B., Wang, Z., & Nakano, K. (2022). Integrated graphical representation of highway scenarios to improve trajectory prediction of surrounding vehicles. *IEEE Transactions on Intelligent Vehicles*, **8**, 1638–1651, https://doi.org/10.1109/tiv.2022.3197179.

Hou, Y., Zhang, X., Cao, X., Lu, Z., & Yuan, X. (2024). Vehicle trajectory prediction model for map-free scenes using the spatio-temporal attentional mechanism. *IEEE Internet of Things Journal*, **12**, 11372–11382. https://doi.org/10.1109/jiot.2024.3506720.

Hu, A., Murez, Z., Mohan, N., Dudas, S., Hawke, J., Badrinarayanan, V., Cipolla, R., & Kendall, A. (2021). Fiery: Future instance prediction in bird's-eye view from surround monocular cameras. In *Proceedings of the IEEE/CVF International Conference on Computer Vision* (pp. 15273–15282). IEEE.

Hu, S., Chen, L., Wu, P., Li, H., Yan, J., & Tao, D. (2022). St-p3: End-to-end vision-based autonomous driving via spatial-temporal feature learning. In *European Conference on Computer Vision* (pp. 533–549). Springer.

Hu, Y., Yang, J., Chen, L., Li, K., Sima, C., Zhu, X., Chai, S., Du, S., Lin, T., Wang, W., Lu, L., Jia, X., Liu, Q., Dai, J., Qiao, Y., & Li, H. (2023). Planning-oriented autonomous driving. In *Proceedings of the IEEE/CVF Conference on Computer Vision and Pattern Recognition* (pp. 17853–17862). IEEE.

Huang, J., Huang, G., Zhu, Z., Ye, Y., & Du, D. (2021). Bevdet: High-performance multi-camera 3d object detection in bird-eye-view. *preprint arXiv:2112.11790*, https://doi.org/10.48550/arxiv.2112.11790.

Hwang, I., & Lee, Y.-G. (2024). The identification of minor impact collisions in a long video for detecting property damages caused by fleeing vehicles using three-dimensional convolutional neural network. *Journal of Computational Design and Engineering*, **11**, 106–121. https://doi.org/10.1093/jcde/qwae016.

Jiang, B., Chen, S., Wang, X., Liao, B., Cheng, T., Chen, J., Zhou, H., Zhang, Q., Liu, W., & Huang, C. (2022). Perceive, interact, predict: Learning dynamic and static clues for end-to-end motion prediction. *preprint arXiv:2212.02181*, https://doi.org/10.48550/arXiv.2212.02181.

Jiang, B., Chen, S., Xu, Q., Liao, B., Chen, J., Zhou, H., Zhang, Q., Liu, W., Huang, C., & Wang, X. (2023). Vad: Vectorized scene representation for efficient autonomous driving. *In Proceedings of the IEEE/CVF International Conference on Computer Vision*, (pp. 8340–8350). IEEE. https://doi.org/10.1109/iccv51070.2023.00766.

Kim, D.-H., & Lee, Y.-G. (2026). End-to-end vehicle lateral control with transformer-based perception and intention-aware view weighting. *Journal of Computational Design and Engineering*, **13**, 1–28. https://doi.org/10.1093/jcde/qwag075.

Lee, S., Park, E., Yi, H., & Lee, S. H. (2020). StRDAN: Synthetic-to-real domain adaptation network for vehicle re-identification. In *Proceedings of the IEEE/CVF Conference on Computer Vision and Pattern Recognition Workshops* (pp. 2590–2597). IEEE.

Lee, S., Woo, T., & Lee, S. H. (2023). Multi-attention-based soft partition network for vehicle re-identification. *Journal of Computational Design and Engineering*, **10**, 488–502. https://doi.org/10.1093/jcde/qwad014.

Li, L., Wang, X., Yang, D., Ju, Y., Zhang, Z., & Lian, J. (2023a). Real-time heterogeneous road-agents trajectory prediction using hierarchical convolutional networks and multi-task learning. *IEEE Transactions on Intelligent Vehicles*, **9**, 4055–4069. https://doi.org/10.1109/tiv.2023.3275164.

Li, P., Ding, S., Chen, X., Hanselmann, N., Cordts, M., & Gall, J. (2023b). Powerbev: A powerful yet lightweight framework for instance prediction in bird's-eye view. *preprint arXiv:2306.10761*, https://doi.org/10.48550/arxiv.2306.10761.

Li, Q., Wang, Y., Wang, Y., & Zhao, H. (2022). Hdmapnet: An online hd map construction and evaluation framework. In *2022 International Conference on Robotics and Automation (ICRA)* (pp. 4628–4634). IEEE.

Li, Y., Ge, Z., Yu, G., Yang, J., Wang, Z., Shi, Y., Sun, J., & Li, Z. (2023c). Bevdepth: Acquisition of reliable depth for multi-view 3d object detection. In *Proceedings of the AAAI Conference on Artificial Intelligence* (vol. **37**, pp. 1477–1485). AAAI Press.

Li, Z., Wang, W., Li, H., Xie, E., Sima, C., Lu, T., Yu, Q., & Dai, J. (2024). Bevformer: Learning bird's-eye-view representation from lidar-camera via spatiotemporal transformers. *IEEE Transactions on Pattern Analysis and Machine Intelligence*, **47**, 2020–2036. https://doi.org/10.1109/TPAMI.2024.3515454.

Liang, M., Yang, B., Hu, R., Chen, Y., Liao, R., Feng, S., & Urtasun, R. (2020). Learning lane graph representations for motion forecasting. In *Computer Vision–ECCV 2020: 16th European Conference, Glasgow, UK, August 23–28, 2020, Proceedings, Part II 16* (pp. 541–556). Springer.

Liao, B., Chen, S., Wang, X., Cheng, T., Zhang, Q., Liu, W., & Huang, C. (2022). Maptr: Structured modeling and learning for online vectorized hd map construction. *preprint arXiv:2208.14437*, https://doi.org/10.48550/arXiv.2208.14437.

Lin, L., Lin, X., Lin, T., Huang, L., Xiong, R., & Wang, Y. (2024a). Eda: Evolving and distinct anchors for multimodal motion prediction. in *Proceedings of the AAAI Conference on Artificial Intelligence*, (vol. **38**, 3432–3440). AAAI Press.

Lin, Z., Liu, Z., Wang, Y., Zhang, L., & Zhu, C. (2024b). Rcbevdet++: toward high-accuracy radar-camera fusion 3d perception network. *preprint arXiv:2409.04979*, https://doi.org/10.48550/arxiv.2409.04979.

Lin, Z., Liu, Z., Xia, Z., Wang, X., Wang, Y., Qi, S., Dong, Y., Dong, N., Zhang, L., & Zhu, C. (2024c). Rcbevdet: Radar-camera fusion in bird's eye view for 3d object detection. In *Proceedings of the IEEE/CVF Conference on Computer Vision and Pattern Recognition* (pp. 14928–14937). IEEE.

Liu, H., Teng, Y., Lu, T., Wang, H., & Wang, L. (2023a). Sparsebev: High-performance sparse 3d object detection from multi-camera videos. In *Proceedings of the IEEE/CVF International Conference on Computer Vision* (pp. 18580–18590). IEEE.

Liu, S., Li, F., Zhang, H., Yang, X., Qi, X., Su, H., Zhu, J., & Zhang, L. (2022). Dab-detr: Dynamic anchor boxes are better queries for detr. *preprint arXiv:2201.12329*, https://doi.org/10.48550/arXiv.2201.12329.

Liu, X., Zou, Y., Che, T., Ding, P., Jia, P., You, J., & Kumar, B. (2019). Conservative wasserstein training for pose estimation. In *Proceedings of the IEEE/CVF International Conference on Computer Vision* (pp. 8262–8272). IEEE.

Liu, Z., Tang, H., Amini, A., Yang, X., Mao, H., Rus, D. L., & Han, S. (2023b). Bevfusion: Multi-task multi-sensor fusion with unified bird's-eye view representation. In *2023 IEEE international conference on robotics and automation (ICRA)* (pp. 2774–2781). IEEE.

Loshchilov, I., & Hutter, F. (2017). Decoupled weight decay regularization. *preprint arXiv:1711.05101*, https://doi.org/10.48550/arXiv.1711.05101.

Lu, J., Zhou, Z., Zhu, X., Xu, H., & Zhang, L. (2022). Learning ego 3d representation as ray tracing. In *European Conference on Computer Vision* (pp. 129–144). Springer.

Luan, B., Xiao, Y., Zhou, W., Zhou, H., Lu, X., & Han, Z. (2026). Intelligent scheduling control for coordinated passing of multiple autonomous trucks at intersections in open-pit coal mines via enhanced multi-agent reinforcement learning. *Journal of Computational Design and Engineering*, **13**, 352–370. https://doi.org/10.1093/jcde/qwaf140.

Luo, W., Park, C., Cornman, A., Sapp, B., & Anguelov, D. (2023). Jfp: Joint future prediction with interactive multi-agent modeling for autonomous driving. In *Conference on Robot Learning* (pp. 1457–1467). PMLR.

Marchetti, F., Becattini, F., Seidenari, L., & Bimbo, A. D. (2020). Mantra: Memory augmented networks for multiple trajectory prediction. In *Proceedings of the IEEE/CVF conference on computer vision and pattern recognition* (pp.7143–7152). IEEE.

Meng, D., Chen, X., Fan, Z., Zeng, G., Li, H., Yuan, Y., Sun, L., & Wang, J. (2021). Conditional detr for fast training convergence. In *Proceedings of the IEEE/CVF international conference on computer vision* (pp. 3651–3660). IEEE.

Nayakanti, N., Al-Rfou, R., Zhou, A., Goel, K., Refaat, K. S., & Sapp, B. (2023). Wayformer: Motion forecasting via simple & efficient attention networks. In *2023 IEEE International Conference on Robotics and Automation (ICRA)* (pp. 2980–2987). IEEE.

Ngiam, J., Caine, B., Vasudevan, V., Zhang, Z., Chiang, H.-T. L., Ling, J., Roelofs, R., Bewley, A., Liu, C., Venugopal, A., Weiss, D., Sapp, B., Chen, Z., & Shlens, J. (2021). Scene transformer: A unified architecture for predicting multiple agent trajectories. *preprint arXiv:2106.08417*, https://doi.org/10.48550/arxiv.2106.08417.

Pei, M., Shi, S., Chen, X., Liu, X., & Shen, S. (2025). Foresight in motion: Reinforcing trajectory prediction with reward heuristics. In *Proceedings of the IEEE/CVF International Conference on Computer Vision* (pp. 28303–28312). IEEE.

Perez, E., Strub, F., De Vries, H., Dumoulin, V., & Courville, A. (2018). Film: Visual reasoning with a general conditioning layer. In *Proceedings of the AAAI conference on artificial intelligence, vol. 32* (pp. 3942–3951). AAAI Press.

Philion, J., & Fidler, S. (2020). Lift, splat, shoot: Encoding images from arbitrary camera rigs by implicitly unprojecting to 3d. In *Computer Vision–ECCV 2020: 16th European Conference, Glasgow, UK, August 23–28, 2020, Proceedings, Part XIV 16* (pp. 194–210). Springer.

Qi, C. R., Su, H., Mo, K., & Guibas, L. J. (2017). Pointnet: Deep learning on point sets for 3d classification and segmentation. In *Proceedings of the IEEE Conference on Computer Vision and Pattern Recognition* (pp. 652–660). IEEE.

Ren, Y., Liu, L., Lan, Z., Cui, Z., & Yu, H. (2025). Mlb-traj: Map-free trajectory prediction with local behavior query for autonomous driving. *IEEE Internet of Things Journal*, **12**, 28556–28571. https://doi.org/10.1109/JIOT.2025.3568029.

Schmidt, J., Jordan, J., Gritschneder, F., & Dietmayer, K. (2022). Crat-pred: Vehicle trajectory prediction with crystal graph convolu-tional neural networks and multi-head self-attention. In *2022 International Conference on Robotics and Automation (ICRA)* (pp. 7799–7805). IEEE.

Schmidt, J., Jordan, J., Gritschneder, F., Monninger, T., & Dietmayer, K. (2023). Exploring navigation maps for learning-based motion prediction. In *2023 IEEE International Conference on Robotics and Automation (ICRA)* (pp. 3539–3545). IEEE.

Shi, S., Jiang, L., Dai, D., & Schiele, B. (2022). Motion transformer with global intention localization and local movement refinement. *Advances in Neural Information Processing Systems*, **35**, 6531–6543. https://doi.org/10.52202/068431-0473.

Sun, J., Yuan, C., Sun, S., Wang, S., Han, Y., Ma, S., Huang, Z., Wong, A., Tee, K. P., & Ang, M. H. (2024a). Controlmtr: Control-guided motion transformer with scene-compliant intention points for feasible motion prediction. In *2024 IEEE 27th International Conference on Intelligent Transportation Systems (ITSC)* (pp. 1507–1514). IEEE.

Sun, W., Lin, X., Shi, Y., Zhang, C., Wu, H., & Zheng, S. (2024b). Sparsedrive: End-to-end autonomous driving via sparse scene representation. *preprint arXiv:2405.19620*, https://doi.org/10.48550/arXiv.2405.19620.

Tabelini, L., Berriel, R., Paixao, T. M., Badue, C., De Souza, A. F., & Oliveira-Santos, T. (2021). Keep your eyes on the lane: Real-time attention-guided lane detection. In *Proceedings of the IEEE/CVF Conference on Computer Vision and Pattern Recognition* (pp. 294–302). IEEE.

Vaswani, A. (2017). Attention is all you need. In *Advances in Neural Information Processing Systems, vol. 30* (pp. 5998–6008).

Wilson, B., Qi, W., Agarwal, T., Lambert, J., Singh, J., Khandelwal, S., Pan, B., Kumar, R., Hartnett, A., Pontes, J. K., Ramanan, D., Carr, P., & Hays, J. (2023). Argoverse 2: Next generation datasets for self-driving perception and forecasting. *preprint arXiv:2301.00493*, https://doi.org/10.48550/arxiv.2301.00493.

Xiang, J., Nan, Z., Song, Z., Huang, J., & Li, L. (2023). Map-free trajectory prediction in traffic with multi-level spatial-temporal modeling. *IEEE Transactions on Intelligent Vehicles*, **9**, 3258–3270. https://doi.org/10.1109/tiv.2023.3342430.

Yang, C., Chen, Y., Tian, H., Tao, C., Zhu, X., Zhang, Z., Huang, G., Li, H., Qiao, Y., Lu, L., Zhou, J., & Dai, J. (2023). Bevformer v2: Adapting modern image backbones to bird's-eye-view recognition via perspective supervision. In *Proceedings of the IEEE/CVF Conference on Computer Vision and Pattern Recognition* (pp. 17830–17839). IEEE.

Ye, M., Cao, T., & Chen, Q. (2021). Tpcn: Temporal point cloud networks for motion forecasting. In *Proceedings of the IEEE/CVF Conference on Computer Vision and Pattern Recognition* (pp. 11318–11327). IEEE.

Ye, T., Jing, W., Hu, C., Huang, S., Gao, L., Li, F., Wang, J., Guo, K., Xiao, W., Mao, W., Zheng, H., Li, K., Chen, J., & Yu, K. (2023). Fusionad: Multi-modality fusion for prediction and planning tasks of autonomous driving. *arXiv preprint arXiv:2308.01006*, https://doi.org/10.48550/arxiv.2308.01006.

Yuan, Y., & Kitani, K. (2020). Dlow: Diversifying latent flows for diverse human motion prediction. In *European Conference on Computer Vision* (pp. 346–364). Springer.

Yuan, Y., Weng, X., Ou, Y., & Kitani, K. M. (2021). Agentformer: Agent-aware transformers for socio-temporal multi-agent forecasting. In *Proceedings of the IEEE/CVF International Conference on Computer Vision* (pp. 9813–9823). IEEE.

Zhang, B., Song, N., & Zhang, L. (2024). Decoupling motion forecasting into directional intentions and dynamic states. *preprint arXiv:2410.05982*, https://doi.org/10.48550/arXiv.2410.05982.

Zhang, H., Li, F., Liu, S., Zhang, L., Su, H., Zhu, J., Ni, L. M., & Shum, H.-Y. (2022a). Dino: Detr with improved denoising anchor boxes for end-to-end object detection. *preprint arXiv:2203.03605*, https://doi.org/10.48550/arxiv.2203.03605.

Zhang, S., Cao, X., Wang, P., & Li, J. (2026). PGEM-DETR: Physics-guided enhancement mechanism for drone-based object detection in adverse visual environments. *Journal of Computational Design and Engineering*, **13**, 60–74. https://doi.org/10.1093/jcde/qwaf121.

Zhang, Y., Zhu, Z., Zheng, W., Huang, J., Huang, G., Zhou, J., & Lu, J. (2022b). Beverse: Unified perception and prediction in birds-eye-view for vision-centric autonomous driving. *arXiv preprint arXiv:2205.09743*, https://doi.org/10.48550/arXiv.2205.09743.

Zhao, H., Gao, J., Lan, T., Sun, C., Sapp, B., Varadarajan, B., Shen, Y., Shen, Y., Chai, Y., Schmid, C., Li, C., & Anguelov, D. (2021). Tnt: Target-driven trajectory prediction. in *Conference on Robot Learning*, pp. 895–904, PMLR.

Zhong, Y., & Lee, S. H. (2025). GazeSymCAT: A symmetric cross-attention transformer for robust gaze estimation under extreme head poses and gaze variations. *Journal of Computational Design and Engineering*, **12**, 115–129. https://doi.org/10.1093/jcde/qwaf017.

Zhou, Z., Zhao, X., Wang, Y., Wang, P., & Foroosh, H. (2022). Centerformer: Center-based transformer for 3d object detection. In *European Conference on Computer Vision* (pp. 496–513). Springer.

Zhou, Z., Wang, J., Li, Y.-H., & Huang, Y.-K. (2023). Query-centric trajectory prediction. In *Proceedings of the IEEE/CVF Conference on Computer Vision and Pattern Recognition* (pp. 17863–17873). IEEE.

Zhu, X., Su, W., Lu, L., Li, B., Wang, X., & Dai, J. (2020). Deformable detr: Deformable transformers for end-to-end object detection. *arXiv:2010.04159*, https://doi.org/10.48550/arXiv.2010.04159.

# Appendix A. Multi-agent scene-level prediction

This appendix provides the supplementary multi-agent scene-level analysis referenced in Section 5. The main paper focuses on target-agent trajectory prediction. The analysis below examines whether independently predicted multimodal trajectories can be assembled into interaction-aware joint future hypotheses. This procedure is not part of the core BEVTraj architecture and is used only as an additional analysis of scene-level behavior.

## A.1. Joint-world construction

Given a scene with multiple agents, we first run BEVTraj independently for each agent and retain the top $K$ trajectory modes. We then construct $S$ joint future hypotheses, or joint worlds, by selecting one mode per agent for each world. A simple strategy is to combine confidence-ranked modes independently across agents. However, this strategy can over-rely on the predicted mode confidence, reduce the diversity of the assembled joint hypotheses, and produce interaction-inconsistent futures because each agent is selected without considering the trajectories already assigned to the same world. Following the general idea of joint future prediction (Luo *et al.*, 2023), we therefore use a lightweight assignment-based procedure.

For each agent $i$, assigning mode $k$ to world $s$ is scored using

$$\begin{aligned} C_i(s,k) &= w_{\text{prob}} c_{\text{prob}}(i,k) + w_{\text{pair}} c_{\text{pair}}(i,k;\mathcal{T}_s) \\ &+ w_{\text{ego}} c_{\text{ego}}(i,k) + w_{\text{reg}} c_{\text{reg}}(i,k), \end{aligned} \tag{A1}$$

where $c_{\text{prob}}(i,k) = -\log p_{i,k}$ penalizes low-confidence modes, and $c_{\text{pair}}(i,k;\mathcal{T}_s)$ penalizes sustained close proximity between the candidate trajectory and trajectories already cached in world $s$. Here, $\mathcal{T}_s$ denotes the set of trajectories already assigned to world $s$. The pairwise term is computed using a distance threshold with an $L$-consecutive criterion, so that short transient proximity is treated differently from sustained overlap. The term $c_{ego}$ is an optional ego-interaction term, and $c_{reg}$ is a lightweight motion regularizer.

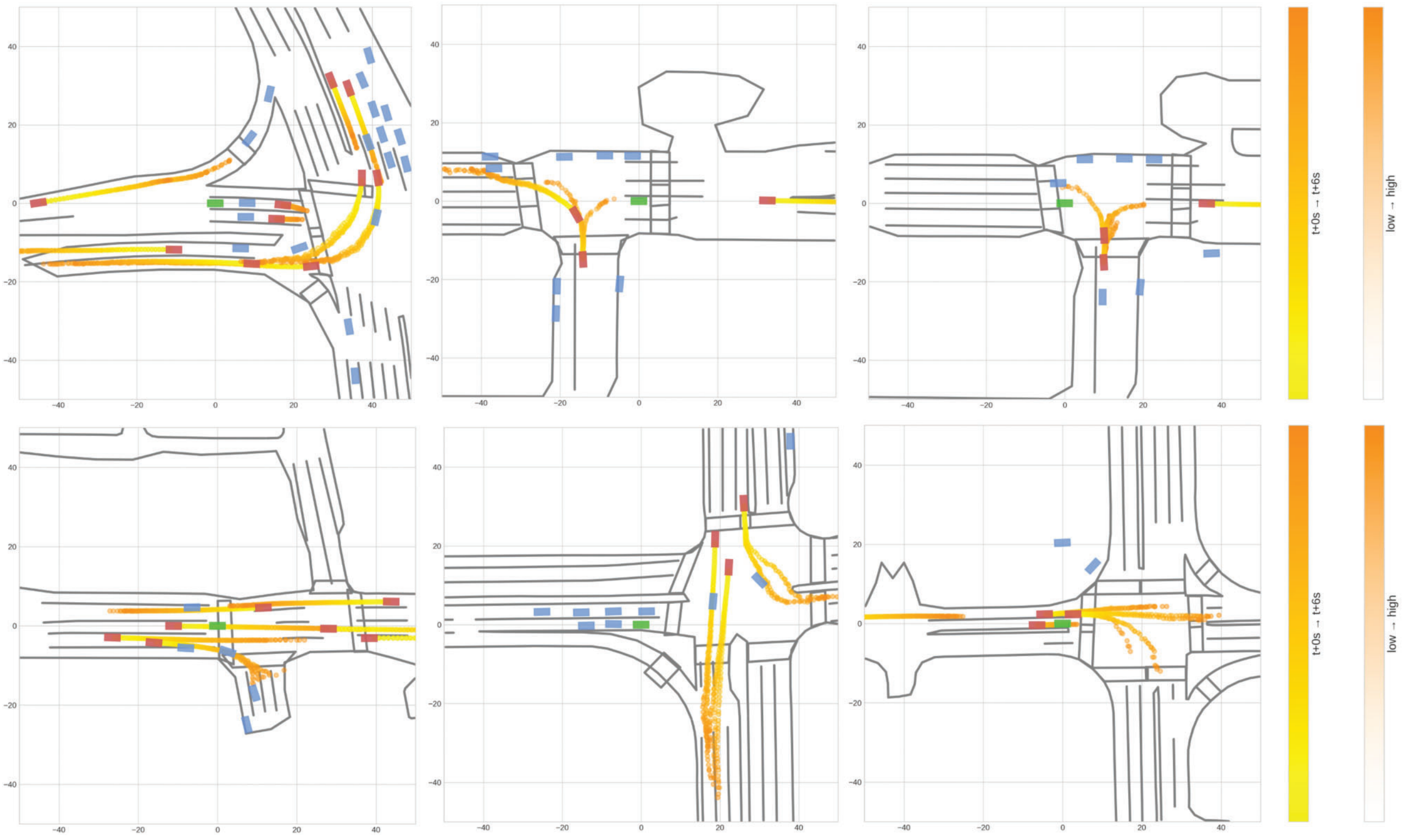

**Figure A1:** Supplementary visualization of BEVTraj predictions for multiple interaction-relevant agents. Red indicates target agents, and green indicates the ego vehicle. For compact presentation, multi-world predictions are aggregated into one visualization, and only the top three modes are shown.

**Table A1:** Supplementary multi-world prediction results on the *nuScenes* validation set.

| Model | AvgMinADE6↓ | AvgMinFDE6↓ | ActorMR↓ | NodeCR↓ | ActorCR↓ |
|---|---|---|---|---|---|
| Autobot | 2.6421 | 6.4028 | 0.7648 | 0.0002 | 0.0022 |
| MTR | 2.1352 | 5.4476 | 0.7229 | 0.0005 | 0.0030 |
| Wayformer | 2.1675 | 5.7544 | 0.7537 | 0.0001 | 0.0018 |
| DEMO | 1.8843 | 4.8902 | 0.6864 | 0.0002 | 0.0025 |
| BEVTraj (Ours) | 1.9285 | 4.9791 | 0.4178 | 0.0000 | 0.0002 |

The assignment is solved with the Hungarian algorithm for each agent, while the trajectory cache $\mathcal{T}_s$ is updated sequentially. This procedure produces a set of joint worlds that accounts for cross-agent compatibility more explicitly than independent confidence-ranked selection, while remaining a post-processing step applied after single-agent multimodal prediction. Figure A1 provides a qualitative visualization of the predicted trajectories for multiple interaction-relevant agents, showing the top three prediction modes.

## A.2. Evaluation metrics and results

We assemble $S = 6$ joint worlds and evaluate them using scene-level metrics adapted from the Argoverse 2 motion-forecasting protocol (Wilson *et al.*, 2023). AvgMinADE6 and AvgMinFDE6 measure the best joint-world displacement error among the six hypotheses. ActorMR measures the fraction of actors whose final displacement error exceeds the miss threshold in the selected joint world. ActorCR measures actor-level collision events by checking whether an actor's predicted trajectory comes within 1.0 m of another actor at any future timestep. NodeCR measures sustained overlap risk at the actor–timestep level using a distance threshold and consecutive-frame criterion; in our experiments, we use $d_{th} = 1.6$ m and $L = 2$.

As shown in Table A1 , the assignment-based construction yields low collision-related rates for BEVTraj in this supplementary evaluation. The result suggests that independently generated multimodal trajectories can be organized into interaction-aware joint hypotheses using a simple post-processing step. However, this analysis should not be interpreted as a replacement for joint multi-agent training or closed-loop planning evaluation. A more complete treatment of scene-level prediction would require explicitly modeling joint distributions during training and evaluating their effect in downstream planning.